\documentclass{article}

\PassOptionsToPackage{numbers}{natbib}

    \usepackage[final]{neurips_2022}

\usepackage[utf8]{inputenc} 
\usepackage[T1]{fontenc}    
\usepackage{hyperref}       
\usepackage{url}            
\usepackage{booktabs}       
\usepackage{amsfonts}       
\usepackage{nicefrac}       
\usepackage{microtype}      
\usepackage{xcolor}         

\usepackage{algorithm}
\usepackage{algpseudocode}
\usepackage[title]{appendix}

\usepackage{amsmath, amsthm, amssymb}

\usepackage{wrapfig}
\renewcommand{\d}[1]{\ensuremath{\operatorname{d}\!{#1}}}

\newcommand\nps[1]{\textcolor{black}{#1}}

\input{shortcuts.tex}
\usepackage{amsmath,amsthm,amsfonts,amssymb,amscd}
\usepackage{enumerate}
\usepackage{fancyhdr}
\usepackage{mathrsfs}
\usepackage{xcolor}
\usepackage{graphicx}
\usepackage{listings}
\usepackage{amsmath}
\usepackage{hyperref}
\usepackage{mathtools}
\usepackage{balance}

\newcommand{\PD}[2]{\frac{\partial{#1}}{\partial{#2}}}

\title{NeuPhysics: Editable Neural Geometry and Physics from Monocular Videos}

\author{%
Yi-Ling Qiao \thanks{Equal contribution.} \\
University of Maryland, College Park
\And
Alexander Gao $^*$ \\
University of Maryland, College Park
\And
Ming C. Lin \\
University of Maryland, College Park
}

\author{%
Yi-Ling Qiao$^{\dagger}$ \ \ \ Alexander Gao$^{\dagger}$\ \ \ Ming C. Lin  \\
\small{University of Maryland, College Park} \\
\small{$^\dagger$Equal Contribution} \\
}

\begin{document}

\maketitle

\begin{abstract}
We present a method for learning 3D geometry and physics parameters of a dynamic scene from only a monocular RGB video input. To decouple the learning of underlying scene geometry from dynamic motion, we represent the scene as a time-invariant signed distance function (SDF) which serves as a reference frame, along with a time-conditioned deformation field. We further bridge this neural geometry representation with a differentiable physics simulator by designing a two-way conversion between the neural field and its corresponding hexahedral mesh, enabling us to estimate physics parameters from the source video by minimizing a cycle consistency loss. Our method also allows a user to interactively edit 3D objects from the source video by modifying the recovered hexahedral mesh, and propagating the operation back to the neural field representation. Experiments show that our method achieves superior mesh and video reconstruction of dynamic scenes compared to competing Neural Field approaches, and we provide extensive examples which demonstrate its ability to extract useful 3D representations from videos captured with consumer-grade cameras.
\end{abstract}

\section{Introduction}
\label{sec:intro}
Monocular RGB videos are ubiquitous today, and computers' ability to understand spatial and physical properties embedded in their contents is essential for many applications. Can you imagine one day being able to build a fully-interactive digital twin of your surrounding world using just a smartphone? You could capture a moving, fully 3D portrait of yourself (Sec.~\ref{sec:video}), estimate the physical material parameters of objects and interact with them virtually (Sec.~\ref{sec:estimate}), playback existing recordings from new angles (Sec.~\ref{sec:novel}), or even edit your favorite cartoon videos (Sec.~\ref{sec:edit}). In this paper, we design a system that makes progress toward realizing all of these things. Taking a single RGB video as input, our method estimates the geometry and dynamics parameters of the scene, parameterized by neural fields~\cite{mildenhall2020nerf}, and enables subsequent editing and physically-based simulation.

While humans can easily infer 3D shapes and physical properties of their environment even from a single eye, these tasks remain challenging for computer algorithms. Geometry reconstruction from a single view is an intrinsically under-constrained problem: there is usually more than one 3D structure that could result in a given 2D projection of a scene. Another key challenge lies in choosing a suitable representation to model the complex geometry of the real world. Explicit representations like surface mesh~\cite{chang2015shapenet,mo2019partnet} are widely used in computer graphics, but they suffer from limited resolution and fixed topology. Implicit representations such as signed distance functions (SDFs) describe complex geometric structure more elegantly, but are not directly suitable for reasoning about dynamics properties. Estimating dynamics parameters is yet more challenging, since it is contingent upon correctly estimating geometry and its variation over time.

In this paper, we design an end-to-end differentiable rendering and simulation pipeline that can directly learn the geometry and dynamics from a monocular video. By assuming the existence of a one-to-one dense mapping between temporal frames in 3D space, our method decouples the geometry estimation task into two complementary parts: a static reference frame, and a mapping function that encodes motion. The static reference frame consists of three neural fields: Appearance, Rigidity, and an SDF for geometry. Given a 3D spatial coordinate as input, these fields respectively output its RGB color, rigidity mask, and signed distance to the nearest surface, in the reference frame. Time-variant motion is encoded in a separate Motion field, which takes as input a 3D spatial coordinate (sampled along a ray), plus a learned latent code denoting a particular discrete time step, and outputs the 3D displacement from the sampled coordinate back to its corresponding reference frame coordinate. We utilize the density function from NeuS~\cite{wang2021neus} for rendering, and the Rigidity mask from Non-Rigid NeRF~\cite{tretschk2021nonrigid} to regularize the motion and obtain foreground/background separation.

A convenient property that emerges from factorizing the scene into static geometry + dynamic motion fields is that by querying the motion field, we can establish a point-wise correspondence across frames, which we leverage to infer dynamics properties of the scene. Specifically, physics parameters are optimized by minimizing the difference between results from the differentiable simulator~\cite{du2021diffpd} and motion field. Upon learning reasonable parameters, the differentiable simulator can be used for physically-based animation and editing of the scene. Experiments show that our method outperforms competing approaches in surface reconstruction, video reconstruction, and novel view synthesis tasks, both qualitatively and quantitatively. Multiple examples also demonstrate our full pipeline's physically-based scene understanding and interactive scene-editing capabilities. Our code and data are available here: \url{https://sites.google.com/view/neuphysics}

In summary, the key contributions of this work are as follows:
\vspace{-\topsep}
\begin{itemize}
\setlength{\itemsep}{1mm}
\setlength{\parskip}{0pt}
    \item \nps{We present a framework for estimating and editing the geometry, appearance, and physics parameters directly from a single monocular video (Figure~\ref{fig:computation}).} 
    \item \nps{By decoupling the learning of static SDF geometry and dynamic motion, our method effectively regularizes this under-constrained problem and enables \textit{reconstruction of high-quality dynamic surface mesh sequences from monocular videos} (Figure~\ref{fig:geometry}).}
    \item \nps{By connecting the neural-fields-based geometry reconstruction to a differentiable physics engine, this approach eliminates the need for manual construction of intermediate 3D models, and avoids poor initialization for the differentiable simulator, empowering users to \textit{estimate the modeling and physics parameters from scratch} (Figure~\ref{fig:estimate}).}
    \item \nps{We design a two-way conversion (Figure~\ref{fig:edit}\&~\ref{fig:computation}) between the explicit hexahedral mesh and implicit neural fields in our end-to-end modeling, simulation and rendering pipeline, enabling users to (1) \textit{extract high-quality explicit meshes} from the implicit fields, and (2) \textit{interact with the implicit 3D field (e.g. add, delete, move, deform, and simulate)} by directly manipulating its explicit mesh counterpart.  }
\end{itemize}

\section{Related Work}
\label{sec:related}
Differentiable volume rendering with implicit neural representations has recently been used to synthesize novel views, estimate geometry and appearance from images and videos. Neural Radiance Fields (NeRF)~\cite{mildenhall2020nerf}, and following works~\cite{dellaert2020neural,sitzmann2020implicit,boss2021nerd,yang2021geometry} have achieved photorealistic rendering results even on complex scenes. The coordinate-based multi-layer perceptron (MLP) representation 
equips the scene with high resolution~\cite{zhang2022iron}, can be effectively regularized~\cite{kaizhang2020}, and is memory-efficient. Compared with differentiable surface rendering~\cite{david2019mitsuba,nimier2020radiative}, volume rendering can backpropagate gradients through the entire space, which leads to better optimization of complex geometry that contains topological changes, and mitigates getting stuck in local minima.
The original density field from NeRF cannot directly reconstruct high-quality surface meshes, so neural SDF fields~\cite{wang2021neus,yariv2021volume,Oechsle2021unisurf,noguchi2022watch,yang2022banmo} have been proposed for multi-view geometry reconstruction. Compared to other discrete representations like meshes~\cite{wang2018pixel2mesh,kato2018neural,liu2019soft}, point clouds~\cite{fan2017point,lin2018learning} and voxels~\cite{choy20163d,xie2019pix2vox}, SDF fields have higher resolution, and do not require pixel-level segmentation~\cite{niemeyer2020differentiable,yariv2020multiview,kellnhofer2021neural} for reconstruction tasks. Current approaches that employ neural SDF fields do not converge to accurate geometry when faced with dynamic scenes, while our method does so from a single video, which we show in figure~\ref{fig:geometry}.

Recent work has extended NeRF to videos of dynamic scenes~\cite{park2021hypernerf,attal2021torf,zhang2021editable}, but they primarily focus on 2D image synthesis, not reconstructing 3D geometry. These works either directly condition the neural fields on time ~\cite{du2021nerflow,gao2021dynamic,xian2021space} or they learn a canonical field separately from time-dependent motion~\cite{park2021nerfies,pumarola2020dnerf,tretschk2021nonrigid,chen2022virtual}. We choose the latter strategy because it better constrains the learning problem based on a reasonable inductive bias, and importantly, it provides us with point-wise correspondence between time steps that is useful for dynamics estimation. By representing geometry with a neural SDF field, and applying strong geometric priors, we can reconstruct dynamic 3D geometry better than existing dynamic extensions of NeRF, while even improving image reconstruction as shown in Section~\ref{sec:video}.

Differentiable simulators enable physics priors to better fit into data-driven approaches, and have therefore increasingly been applied to robotics and computer graphics tasks~\cite{Learning2020Song,du2020stokes,heiden2021neuralsim,li2021soft,Spielberg2019,xu2021accelerated}. These physically-based simulation methods have been developed for dynamical systems with different properties, including rigid bodies~\cite{Belbute2018,Qiao2020Scalable,Degrave2019,brax2021github,ma2022risp}, soft bodies~\cite{Hu2019:ICLR,Hu2019:ICRA,Qiao2021Differentiable,Geilinger2020add,du2021diffpd,li2022contact,murthy2021gradsim,heiden2021disect}, articulated bodies~\cite{ha2017joint,werling2021fast,qiao2021Efficient}, cloth~\cite{li2022diffcloth,Liang2019}, and fluids~\cite{Um2020solver,wandel2021learning,Holl2020,takahashi2021differentiable}. 

\section{Methods}
\label{sec:method}
\begin{figure}
    \centering
    \includegraphics[width=1\linewidth]{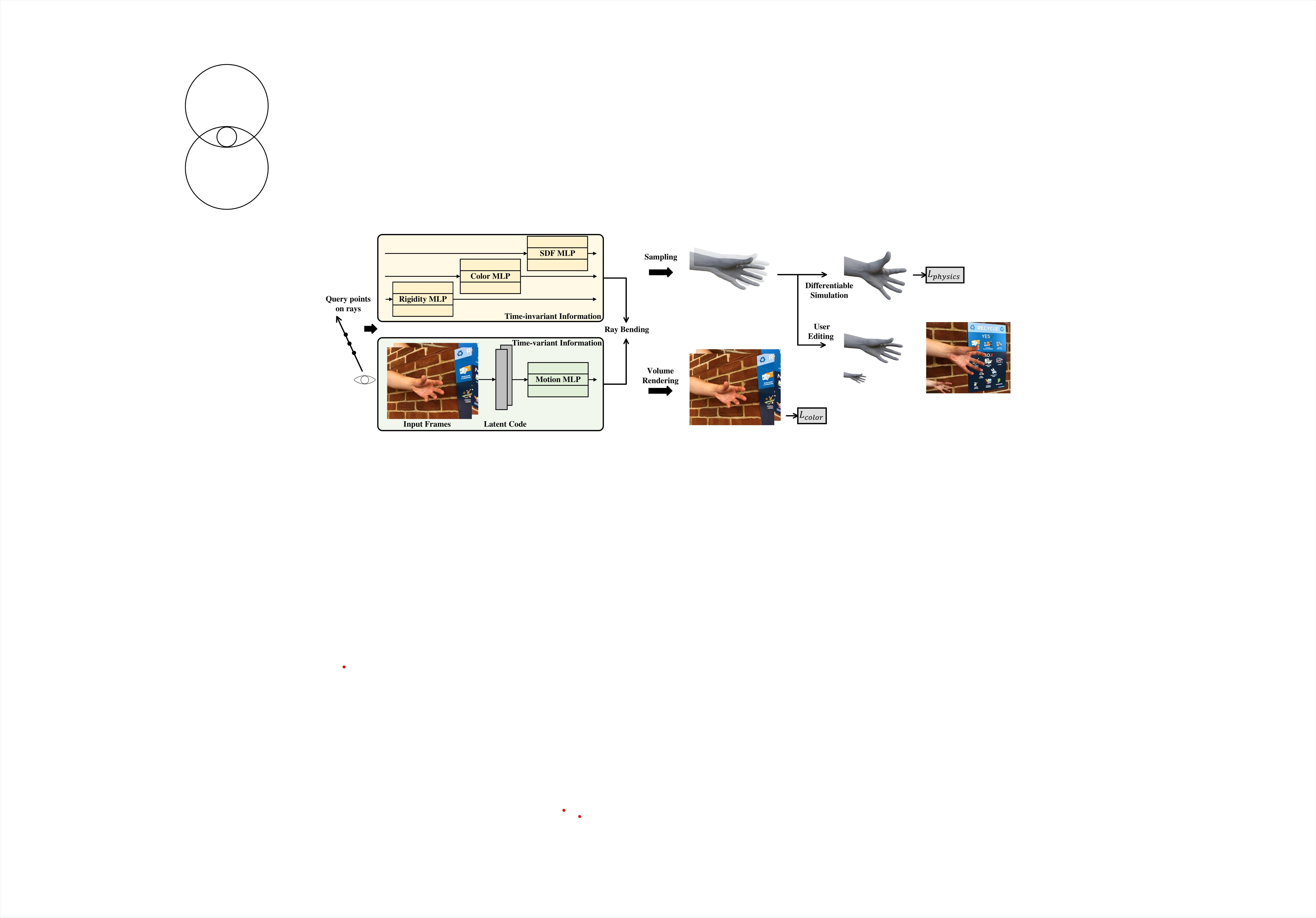}
    \vspace*{-1em}
    \caption{{\bf Overall workflow}. {\em NeuPhysics} takes a monocular video as supervision. The time-invariant fields are defined on a reference frame, while time-dependent motion is applied to each point sampled in space. We then compute color loss from the volume rendered images. A differentiable simulator is embedded after optimizing the neural fields, to learn dynamics parameters and allow scene editing. A detailed computational graph can be found in Figure~\ref{fig:computation}. }
    \label{fig:workflow}
    \vspace*{0em}
\end{figure}

Given a monocular video consisting of $N$ RGB image frames $\{\II_i\}_{i=1}^{N}$, our approach estimates the 3D geometry within the camera's field of view, point-wise correspondence of the dynamic 3D scene across time, and static/dynamic segmentation. By integrating the differentiable physics module, we can learn physics parameters from the dynamic motion sequence, which we leverage for video editing and forecasting. Figure~\ref{fig:workflow} shows an overview of our paper.

\subsection{Rendering Pipeline}
\mypara{SDF and color networks.} Similar to recent Neural Fields methods~\cite{mildenhall2020nerf}, we parameterize the geometry and appearance of the scene using coordinated-based Multi Layer Perceptrons (MLPs), which can render 2D images using the volume rendering scheme proposed in NeuS~\cite{wang2021neus}. Geometry is represented by an SDF network $s_{\theta_{SDF}}(\pp): \mathbb{R}^3 \rightarrow \mathbb{R}$ that maps a spatial coordinate $\pp \in \mathbb{R}^3$ to its signed distance to the nearest surface. Appearance is represented by another MLP $\cc_{\theta_{color}}(\pp): \mathbb{R}^3 \rightarrow \mathbb{R}^3$ that describes the RGB value of each spatial point. During rendering, the points along a ray cast from a given pixel can be written as $\{\pp(t) =\oo + t\vv | t \ge  0\}$, where $\oo$ is the camera center, $\vv$ is the unit direction of the ray, \nps{and $t$ is distance from the camera center}. The pixel's color is then calculated by integrating along the ray:
\begin{equation}
\label{eq:rendering}
    \CC(\oo, \vv) = \int_0^{+\infty} w(s(\pp(t)))\cdot \cc(\pp(t)) \d t,
\end{equation}
where $w(\cdot)$~\cite{wang2021neus} is a weight function describing the contribution of each point's color based on its SDF value. More details can be found in the Appendix.

\mypara{Motion network.} However, the above framework is only designed to handle multi-view images of a static scene, which motivates another component to accommodate temporal information. Instead of directly conditioning the existing networks upon time, we choose to decouple the dynamic and static scene components. Therefore, the SDF and color networks are defined on a static reference frame, while a motion network $\bb_{\theta_{motion}}(\ll_i, \pp): \mathbb{R}^{d_l}\times \mathbb{R}^3\rightarrow \mathbb{R}^3$ estimates the offset field~\cite{tretschk2021nonrigid} for each time step, where $\ll_i\in\mathbb{R}^{d_l}$ is a learned latent feature encoding temporal information about the $i$-th frame $\II_i$. The motion network bends each query ray $\{\pp(t)\}$ to $\{\pp'(t)=\pp(t)+\bb(\ll_i, \pp(t))\}$, which is then queried on the reference frame for SDF and color values.

\mypara{Rigidity network.} Since we want to segment and simulate dynamic objects within the scene downstream, we apply a rigidity network~\cite{tretschk2021nonrigid} $r_{\theta_{rigidity}}(\pp):\mathbb{R}^3 \rightarrow \mathbb{R}$ to distinguish dynamic foreground from static background. The mask $r(\pp)\in [0, 1]$ defined on the reference frame indicates the likelihood that a point belongs to the swept volume of moving elements in the scene. The offset value becomes $\bb_{\theta_{motion}}^{\theta_{rigidity}}(\ll_i, \pp(t)) = r_{\theta_{rigidity}}(\pp)\cdot \bb_{\theta_{motion}}(\ll_i, \pp(t))$. This formulation pushes dynamic regions toward larger $r(\cdot)$ values during training, while driving static areas toward zero.

\subsection{Simulation Pipeline}
\label{sec:simulation}
\begin{wrapfigure}{r}{0.5\linewidth}
\centering
\includegraphics[width=1\linewidth]{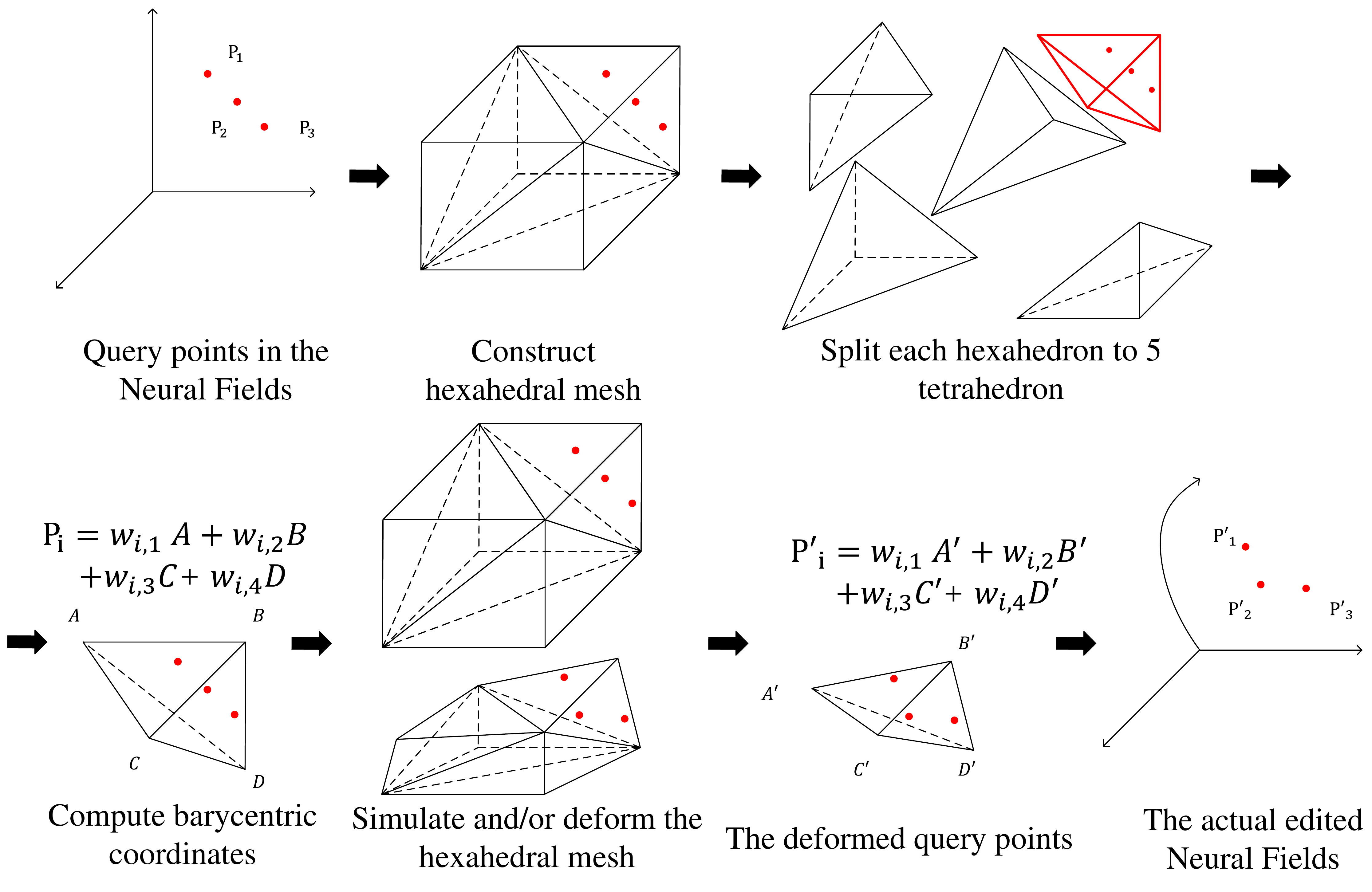}
\vspace{-2em}
\caption{{\bf The editing pipeline}. Users can edit the neural fields by operating on a hexahedral mesh.}
\label{fig:edit}
\vspace*{-1em}
\end{wrapfigure}

\mypara{Geometry reconstruction.} Given the signed distance field $s(\cdot)$ and rigidity mask $r(\cdot)$, we can reconstruct the geometry of the dynamic and static parts of the scene. For example, the foreground in the $i$-th frame is contained within the swept volume of moving objects described by:

\begin{equation}
\label{eq:foreground}
\mathcal{A}_i = \{\pp | s(\pp+\bb(\ll_i, \pp)) \le k_{sdf}, r(\pp+\bb(\ll_i, \pp))\ge k_{rigidity}\},
\end{equation}
where $k_{sdf}$ and $k_{rigidity}$ are constant thresholds for the SDF and rigidity values, respectively. This formulation is amenable to conversion into an explicit mesh representation. \nps{For rendering purposes, a \textit{surface mesh} could be obtained using the marching cubes algorithm~\cite{lorensen1987marching}. Alternatively, we represent the dynamic elements as a \textit{volume mesh}. We sample over a discrete 3D grid to find vertices within volume $\mathcal{A}_i$, and if a grid point $\pp$ satisfies Equation~\ref{eq:foreground}, we add that voxel to a hexahedral volume mesh $\mm_i$. We choose hexahedral rather than tetrahedral volume mesh due to its simplicity and regular structure, and because hexahedral mesh can be simulated by DiffPD~\cite{du2021diffpd}, a differentiable FEM simulator.}
Figure~\ref{fig:edit} visualizes the conversion between the implicit and explicit representation. Triangle surface mesh and tetrahedral volume meshes could be explored in future work.

\mypara{Physics Engine.} From the input video of a dynamic scene, we can extract a mesh sequence $\{\mm_i\}_{i=1}^N$ with consecutive motion. An individual mesh model $\mm_i$ is described by its vertices and hexahedral elements $\{\QQ_i\in\mathbb{R}^{n_v\times 3}, \EE_i\in\mathbb{N}^{n_e\times 8}\}$, where the rows of $\QQ_i$ are the 3D coordinates of $n_v$ vertices, and each row in $\EE_i$ records the indices of eight vertices that form a hexahedron. We focus on modeling dynamic objects as soft bodies in this work. Compared to rigid bodies, soft bodies offer more expressiveness to model a wide range of deformable objects that that are important in applications like physically-based mixed reality and digital twins. As our physics engine, we use Projective Dynamics~\cite{bouaziz2014projective} implemented by DiffPD~\cite{du2021diffpd}. The governing equation can be written as

\begin{equation}
\label{eq:dynamics}
    \MM(\QQ_{i+1}-\QQ_i-h\Dot{\QQ}_i)=h^2(\nabla E(\QQ_{i+1}, \theta_{physics})+\ff_{ext}),
\end{equation}
where $\MM$ is the mass matrix, $\QQ_i$ is the vertex locations at frame $i$, $h$ is the time step, $\Dot{\QQ}_i$ is the velocity, $E$ is the potential energy due to deformation, $\theta_{physics}$ are the physics parameters, and $\ff_{ext}$ are the external forces. Note that the gradient of the potential energy $\nabla E$ can be interpreted as the internal forces within the soft materials. This system uses implicit time integration for increased numerical stability, thus $\nabla E$ is defined on the frame $i+1$. Objects are modeled by homogeneous co-rotated materials, where the material parameters $\theta_{physics}$ include Young’s modulus and Poisson’s ratio. A detailed formulation of the elastic energy $E$ can be found in the supplementary materials of \cite{du2021diffpd}.

System identification can be valuable for constructing faithful digital twins, producing plausible animations, and control of dynamical systems, but estimating states $\QQ$ and $\Dot{\QQ}$, and parameters $\theta_{physics}$ and $\ff_{ext}$ from only a monocular video is challenging due to the high-dimensionality and nonlinearity of the system. Recent progress in differentiable physics tackles this challenge by making the entire simulation differentiable. Gradients $\PD{\mathcal{L}}{\QQ_i}$, $\PD{\mathcal{L}}{\Dot{\QQ}_i}$, $\PD{\mathcal{L}}{\theta_{physics}}$, and $\PD{\mathcal{L}}{\ff_{ext}}$, where $\mathcal{L}$ is a user-defined scalar loss function, are all available, and allow us to estimate the parameters using a gradient-based optimizer like Adam~\cite{kingma2014adam}.

\mypara{Parameter Estimation.}
Running one simulation step forward in time from the $i$-th frame gives us an updated model at frame $i+1$, \nps{dependent on parameters $\theta_{physics}$:\ \ $\mm_{i+1}'(\theta_{physics}) = sim(\theta_{physics}, \mm_i,i\rightarrow i+1)$}. By simulating the reconstructed meshes in the differentiable physics engine $sim(\theta_{physics}, \cdot)$ and comparing with the ground truth reconstruction results as a supervision signal, we optimize the physics parameters $\theta_{physics}$ using gradient-based methods.

One idea for a suitable loss function is to use a metric like Chamfer Distance ${distance}(\mm_{i+1}, \mm_{i+1}'(\theta_{physics}))$ between the extracted and simulated mesh $\mm_{i+1}$ and $\mm_{i+1}'(\theta_{physics})$, searching for the $\theta_{physics}$ that minimizes the distance. However, this strategy requires sampling and mesh reconstruction for every time step, which is computationally expensive. \nps{An ablation study and profiling numbers can be found in Figure~\ref{fig:physics}.} Additionally, the reconstructed mesh might vary significantly across time steps, thus cannot guarantee point-wise correspondence, making the dynamics discontinuous and non-differentiable w.r.t. the networks. 

To save sampling time and make the gradient well-defined, we design a cycle-consistency loss to compute the distance between the simulation results and neural fields. \nps{Starting from an arbitrary mesh $\mm_{i}$ sampled at the $i$-th frame, we run the differentiable simulation for $j$ steps and get $\mm'_{i+j}(\theta_{physics})=sim(\theta_{physics}, \mm_i,i\rightarrow i+j)$. When $\mm'_{i+j}(\theta_{physics})$ is deformed back to the reference frame using the learned motion field, the result should correspond to the original shape $\mm_{i}$. Since $\mm'_{i+j}(\theta_{physics})$ and $\mm_{i}$ share a common topology, an exact point-wise correspondence exists, and we can directly subtract their point clouds to obtain their distance:}

\nps{
\begin{small}
\begin{equation}
\label{eq:cycle}
\mathcal{L}_{physics}=\norm{(\mm_i+\bb_{\theta_{motion}}^{\theta_{rigidity}}(\ll_i, \mm_i))-(\mm'_{i+j}(\theta_{physics})+\bb_{\theta_{motion}}^{\theta_{rigidity}}(\ll_{i+j}, \mm'_{i+j}(\theta_{physics})))}_2
\end{equation}
\end{small}
\vspace*{-1em}
}
\\\\
\noindent
\nps{where $\bb(\ll_i, \mm_i)$ is the motion field at frame $i$ that maps mesh $\mm_i$ back to the reference frame. $\mathcal{L}_{physics}$ is differentiable w.r.t. the motion network, rigidity network, and physics parameters}. Importantly, independent of the simulation duration, we only need to sample $\mm_i$ once, and it can be used as a reference point to compare all other frames.

\mypara{Scene Editing.} The simulator in our pipeline imposes a strong physics prior for forecasting motion and modifying dynamics. Figure~\ref{fig:edit} shows how we edit the signed distance and color fields through the hexahedral mesh. Based on hexahedral mesh $\mm_i$ extracted from the $i$-th frame, we can simulate it to get a modified state $\mm_{i+1}'$. In order to render the modified scene, we add an additional `bending' layer $\bb_{edit}(\mm_i',\cdot)$ on top of the primary motion field $\bb(\ll_i, \cdot)$. 

\nps{When querying a ray point $\pp'$ to render the modified scene, we first check if the point is displaced by our editing operation, i.e. if it is located within the hexahedral mesh $\mm_i'$. The inside-outside-test would be trivial if all hexahedra in $\mm_i'$ remained in a regular grid}, however, they may have deformed during simulation. Thus, we split each hexahedron into five tetrahedra -- a query point $\pp'$ is inside the hexahedron if and only if it is inside one of the tetrahedra. Following \cite{Yuan22NeRFEditing}, we decide whether $\pp'$ is inside a tetrahedron by its barycentric coordinates. For a tetrahedron consisting of four vertices $\{\pp'_i=(x'_i, y'_i, z'_i)\trans\}_{i=1}^4$, the barycentric coordinates $\{c_i\}_{i=1}^4$ of $\pp'=(x', y', z')\trans$ can be computed as $c_i = Det_i / Det_0$, where:
\begin{equation}
Det_0=
\begin{vmatrix}
x'_1 & y'_1 & z'_1 & 1 \\ 
x'_2 & y'_2 & z'_2 & 1 \\ 
x'_3 & y'_3 & z'_3 & 1 \\ 
x'_4 & y'_4 & z'_4 & 1 
\end{vmatrix},
Det_1=
\begin{vmatrix}
x' & y' & z' & 1 \\ 
x'_2 & y'_2 & z'_2 & 1 \\ 
x'_3 & y'_3 & z'_3 & 1 \\ 
x'_4 & y'_4 & z'_4 & 1 
\end{vmatrix}, 
Det_2=
\begin{vmatrix}
x'_1 & y'_1 & z'_1 & 1 \\ 
x' & y' & z' & 1 \\ 
x'_3 & y'_3 & z'_3 & 1 \\ 
x'_4 & y'_4 & z'_4 & 1 
\end{vmatrix}, ...,
\end{equation}

and $Det_i$ is obtained by substituting the $i$-th row of $Det_0$ with $(x', y', z', 1)$. $\pp'$ is inside or on the surface of this tetrahedron if and only if $0\le c_i \le 1$ holds for $i=1,2,3,4$. To determine tetrahedra to check for a given point, we use k-nearest neighbors $\QQ'_{\pp'}=KNN(\mm_i', \pp')$ to locate mesh vertices $\QQ'_{\pp'}$ nearest to query point $\pp$, and select tetrahedra that contain those vertices.

If $\pp'$ does belong to a tetrahedron, then it is displaced along with the mesh. Assuming the modification can be linearly interpolated inside a tetrahedron, the original coordinate $\pp$ in the $i$-th frame becomes the weighted sum $\pp=\sum_{i=1}^4 c_i\pp_i$, where $\pp$ is the reference coordinate corresponding to $\pp'_i$ (before editing). To account for the area that was once occupied by $\mm_i$ but is now empty after editing: we simply assign those points a large positive SDF value such that they are 
effectively transparent during volume rendering and do not influence 3D reconstruction.

\begin{figure}
    \centering
    \includegraphics[width=1\linewidth]{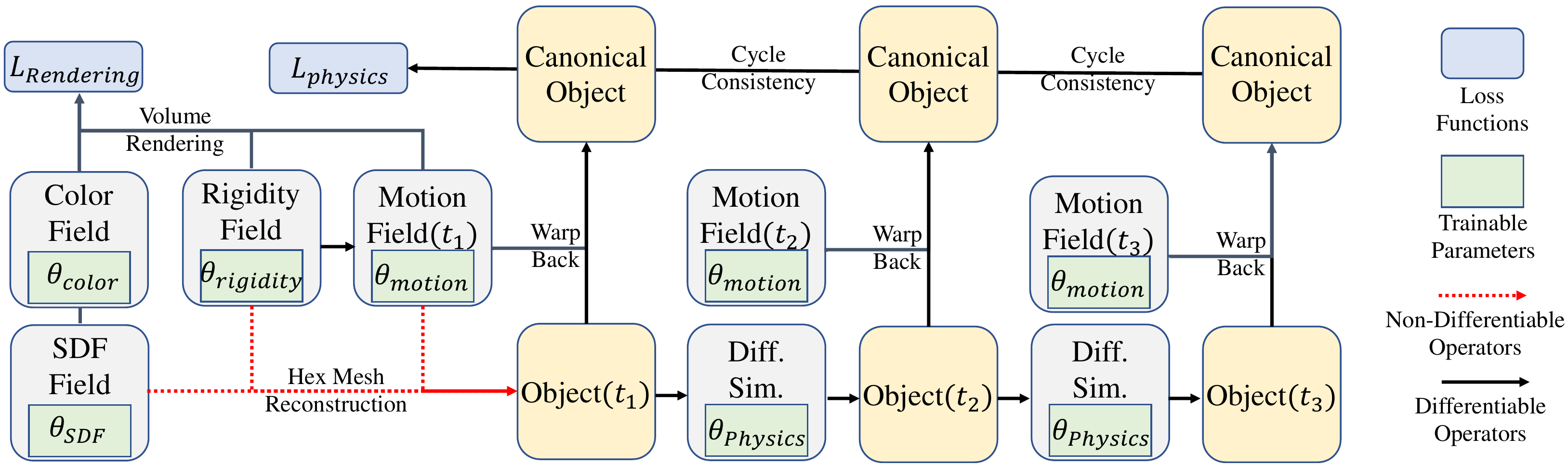}
    \vspace*{-1em}
    \caption{\nps{{\bf Computational Graph During Training}. We show how data transfers among modules in a basic 3-step dynamical systems. All operators are differentiable, except the hex mesh reconstruction for the first frame. In our designed cycle consistency mechanism, the rigidity and motion networks serve as the bridge between the rendering and physics module.}}
    \label{fig:computation}
    \vspace{0em}
\end{figure}

\subsection{Training and Losses}
\label{sec:training}
With video as the main supervision signal used to train our system, we wish to minimize the distance between rendered and the ground truth images $\mathcal{L}_{color}=\norm{\II_i-\hat{\II}_i}_1$. We include additional terms to regularize training. First, an important property of an SDF is its unit norm. Eikonal term~\cite{gropp2020implicit}, $\mathcal{L}_{Eikonal}=\sum_{i,k}(\norm{\pp_{i,k}}_2-1)^2$ is therefore added to the loss. We also want to constrain the motion of our scene (i.e. to be locally rigid, incompressible, and contain mostly static ragions), so we include an offset loss $\mathcal{L}_{offset}=\sum_{i,k}\norm{\bb(\ll_i,\pp_{i,k})}_2+r(\pp_{i,k})$ and divergence loss~\cite{tretschk2021nonrigid} $L_{divergence}=\sum_{i,k}\norm{\text{div}(\bb(\ll_i, \pp_{i,k}))}_2$. Finally, we use the physics loss $\mathcal{L}_{physics}$ when we train for physics parameters. The total loss function is:

\begin{equation}
    \mathcal{L}_{total} = \mathcal{L}_{color} + w_{Eik}\mathcal{L}_{Eikonal} + w_{off}\mathcal{L}_{offset} + w_{div}\mathcal{L}_{divergence} + w_{phy}\mathcal{L}_{physics}
\end{equation}

By default, we set $w_{Eik}=0.1$, $w_{off}=50.0$, and $w_{div}=2.0$, though these weights may be further tuned depending on the scene. \nps{Figure~\ref{fig:computation} demonstrates the data and flow of gradients during training. In our proposed cycle-consistency physics loss, the rendering and simulation modules are connected by the rigidity and motion networks. Instead of jointly training all the parameters together, we adopt a sequential training strategy where the rendering networks are first trained to optimize $\mathcal{L}_{total}$ for 300,000 epochs, and the physics parameters are then trained to optimize $\mathcal{L}_{physics}$ for 100 epochs.} The physics module is trained independently after the geometry networks converge because the differentiable simulation is much more expensive than querying of MLPs. \nps{An ablation study for the efficiency of our training strategy is conducted in Section~\ref{sec:ablation}.} In future work it would be interesting to try a faster simulation engine which could be incorporated as a physics prior \nps{during training of the rendering networks}. A full training cycle typically takes around 18 hours on an NVIDIA A5000 GPU. For the training data (used with consent), we note that the background must contain sufficient detail, or the Structure from Motion implementation by COLMAP~\cite{schoenberger2016sfm,schoenberger2016mvs} could fail.



\section{Experiments}

\label{sec:experiment}

\begin{table*}
\newcolumntype{Z}{S[table-format=2.3,table-auto-round]}
\centering
\setlength{\tabcolsep}{3mm}
\ra{1.05}
\small
\resizebox{1\linewidth}{!}{
\begin{tabular}{@{}lccccccccccccc@{}}
  \toprule
  \multirow{2}[3]{*}{Method}  & \multicolumn{3}{c}{Sitting}  &  \multicolumn{3}{c}{Standing}   &  \multicolumn{3}{c}{Basketball}  &  \multicolumn{3}{c}{Hand}  \\
  \cmidrule(l{3mm}r{3mm}){2-4} \cmidrule(l{3mm}r{3mm}){5-7} \cmidrule(l{3mm}r{3mm}){8-10} \cmidrule(l{3mm}r{3mm}){11-13}
  & LPIPS$\downarrow$ & SSIM$\uparrow$ & PSNR$\uparrow$& LPIPS$\downarrow$ & SSIM$\uparrow$ & PSNR$\uparrow$& LPIPS$\downarrow$ & SSIM$\uparrow$ & PSNR$\uparrow$& LPIPS$\downarrow$ & SSIM$\uparrow$ & PSNR$\uparrow$   \\
  \midrule
  D-NeRF   & \textbf{0.163}  & 0.914  & 34.217 &  0.251 & 0.868 & \textbf{33.036} &  0.332 & 0.831 & 32.803 &  \textbf{0.165} & 0.885 & 34.146\\
  NeuS   & 0.239  & 0.911  & \textbf{34.563} &  0.301 & 0.829 & 32.888 &  0.330 & 0.862 & 34.440 &  0.305 & 0.908 & 34.485\\
  NRNeRF   & 0.223  & 0.892  & 32.019 &  0.269 & 0.863 & 31.549 &  0.340 & 0.830 & 32.411 &  0.205 & 0.913 & 33.999\\
  Ours   & 0.166  & \textbf{0.937} & \textbf{34.563} &  \textbf{0.250} & \textbf{0.869} & 32.881 &  \textbf{0.235} & \textbf{0.912} & \textbf{35.974} &  0.225 & \textbf{0.915} & \textbf{34.710}
  \\ \bottomrule
\end{tabular}
}
\vspace*{-0.5em}
\caption{{\bf Quantitative evaluation of video reconstruction}. We evaluate our method on both real-world and synthetic videos, outperforming alternative methods in most cases.}
\label{tab:recon}
\vspace{-0.5em}
\end{table*}

\vspace{-0.5em}
\subsection{Dynamic Geometry Reconstruction}
\label{sec:video}
\vspace{-0.5em}

\begin{figure}
\centering
\begin{tabular}{@{}c@{\hspace{0.3mm}}c@{\hspace{0.3mm}}c@{\hspace{0.3mm}}c@{\hspace{0.3mm}}c@{\hspace{0.3mm}}c@{\hspace{0.3mm}}c@{\hspace{0.3mm}}c@{\hspace{0.3mm}}@{}}
    \includegraphics[width=0.12\linewidth]{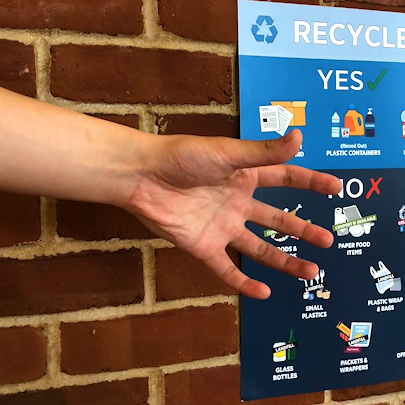} &
    \includegraphics[width=0.12\linewidth]{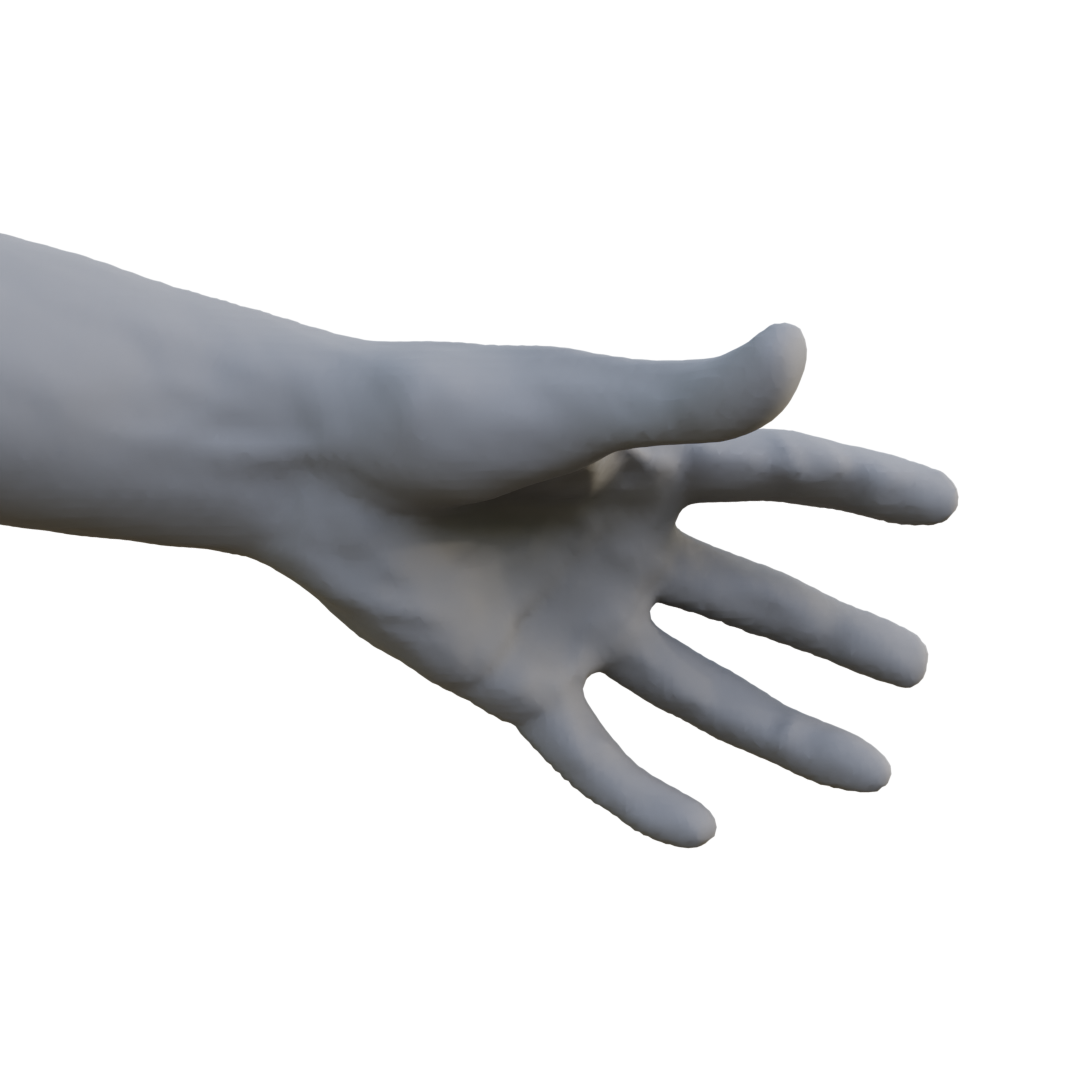} &
    \includegraphics[width=0.12\linewidth]{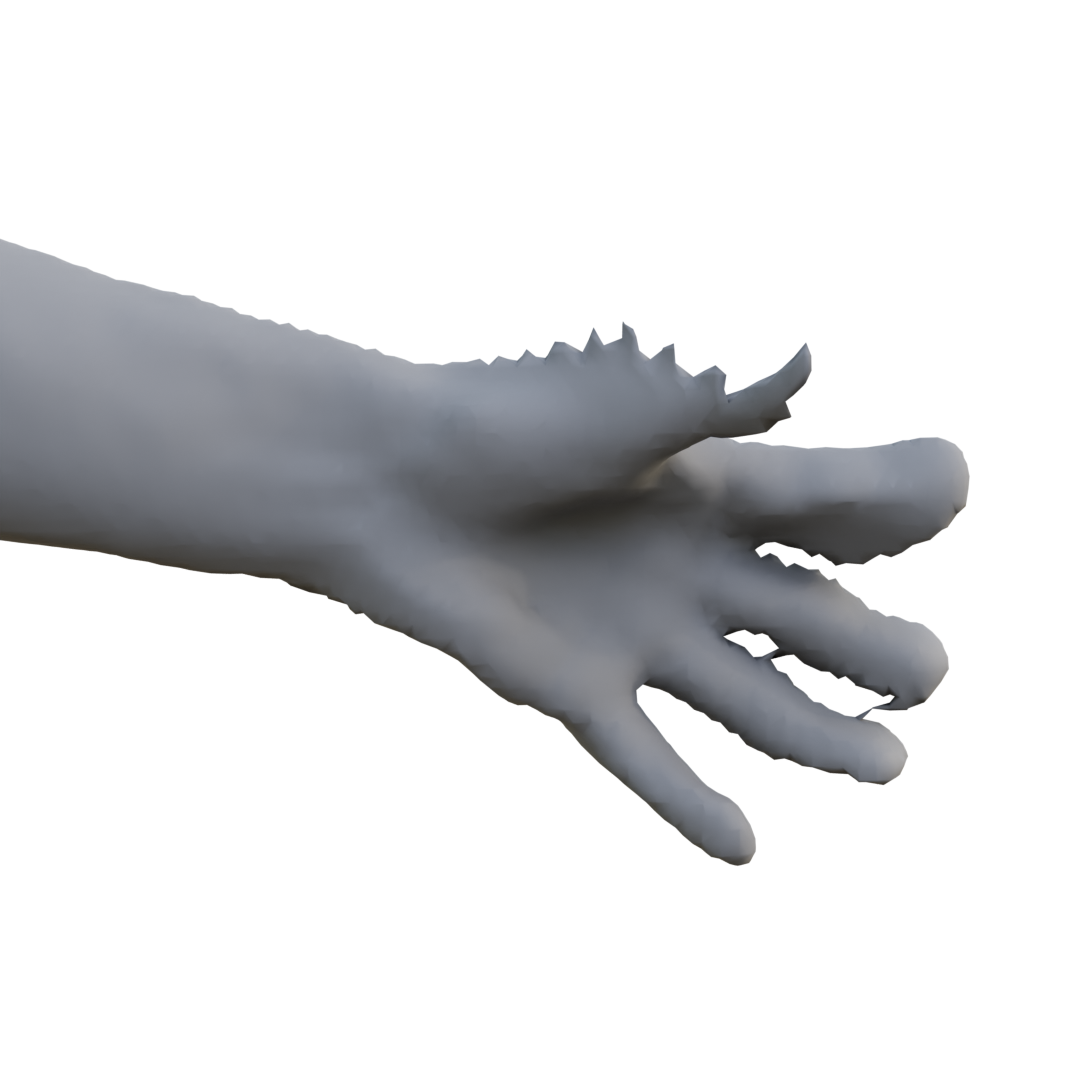} &
    \includegraphics[width=0.12\linewidth]{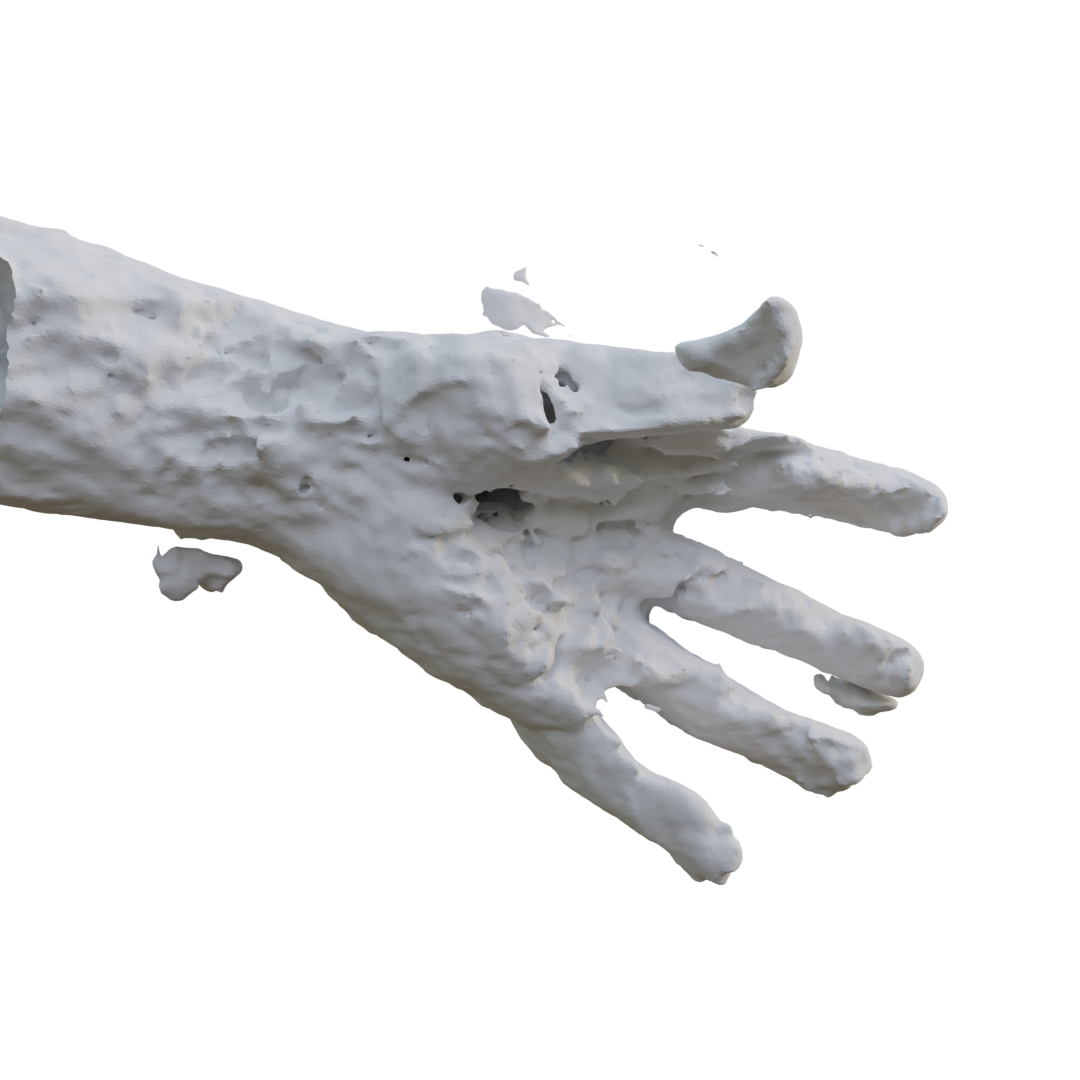} &
    \includegraphics[width=0.12\linewidth]{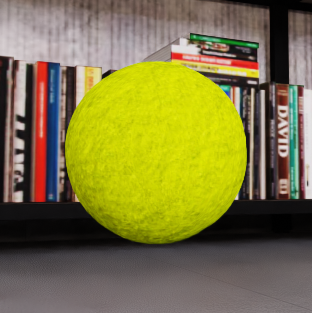} &
    \includegraphics[width=0.12\linewidth]{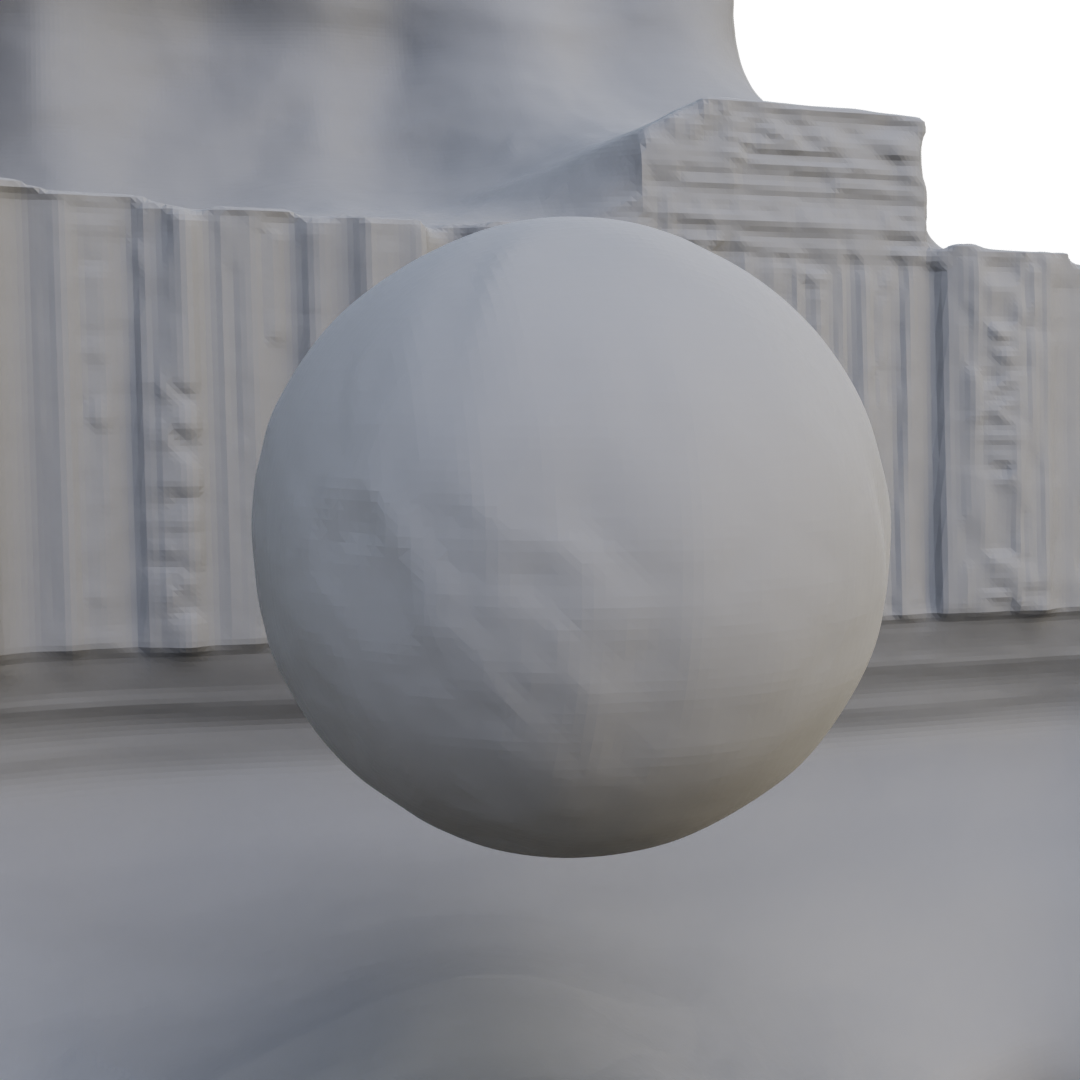}  &
    \includegraphics[width=0.12\linewidth]{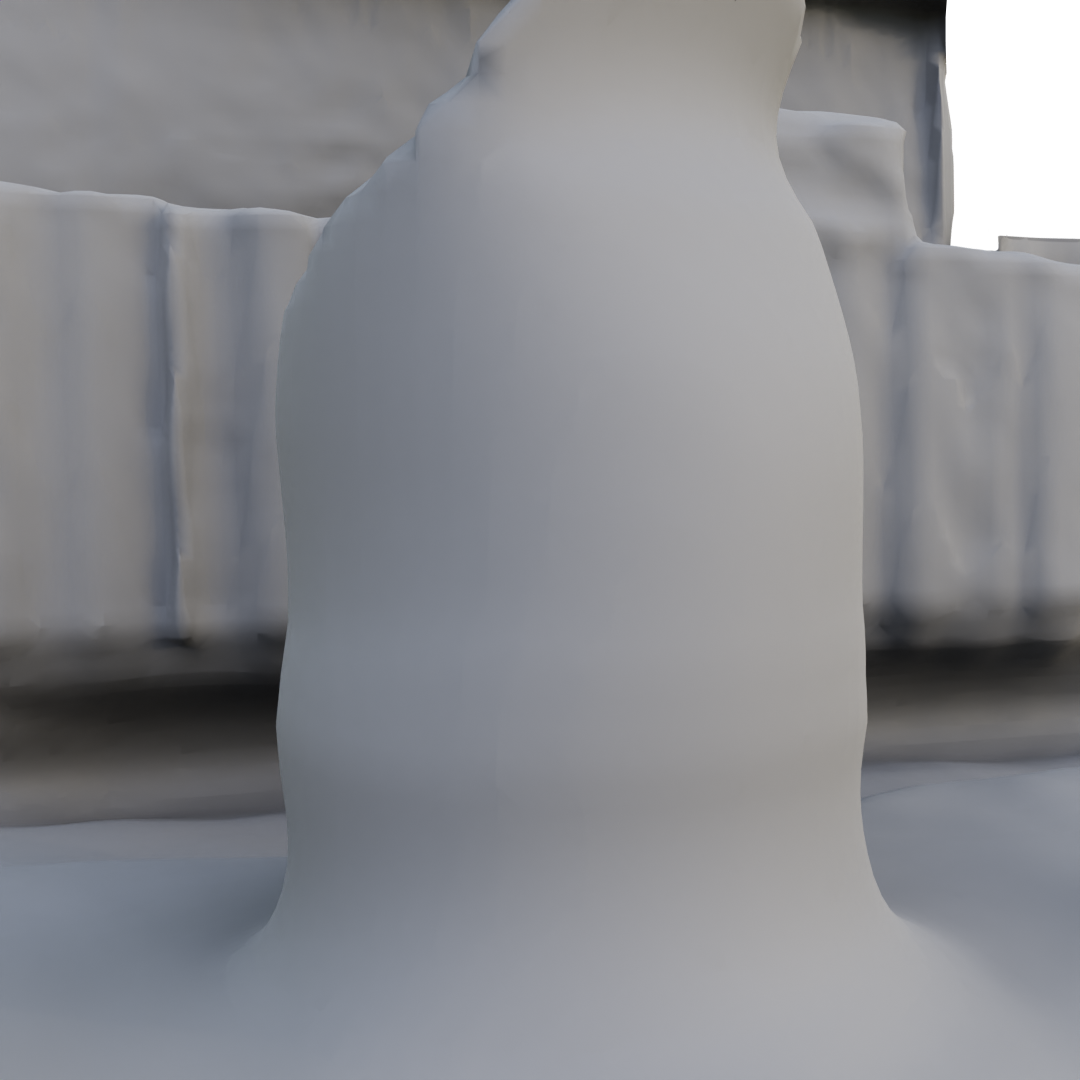}  &
    \includegraphics[width=0.12\linewidth]{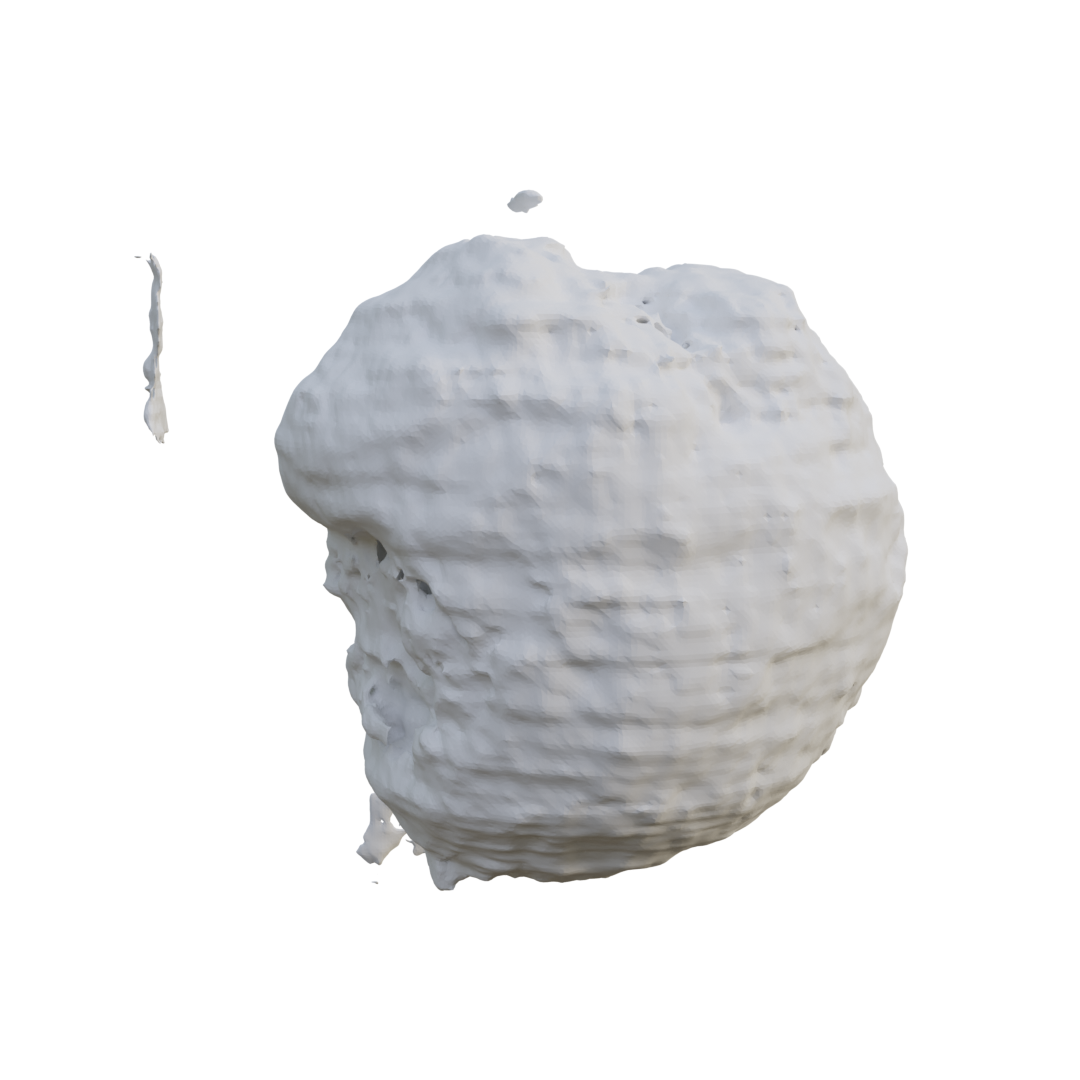} \\
    \includegraphics[width=0.12\linewidth]{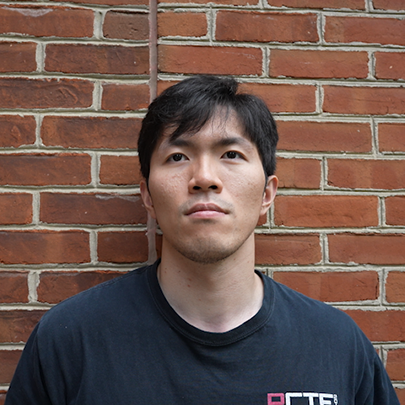} &
    \includegraphics[width=0.12\linewidth]{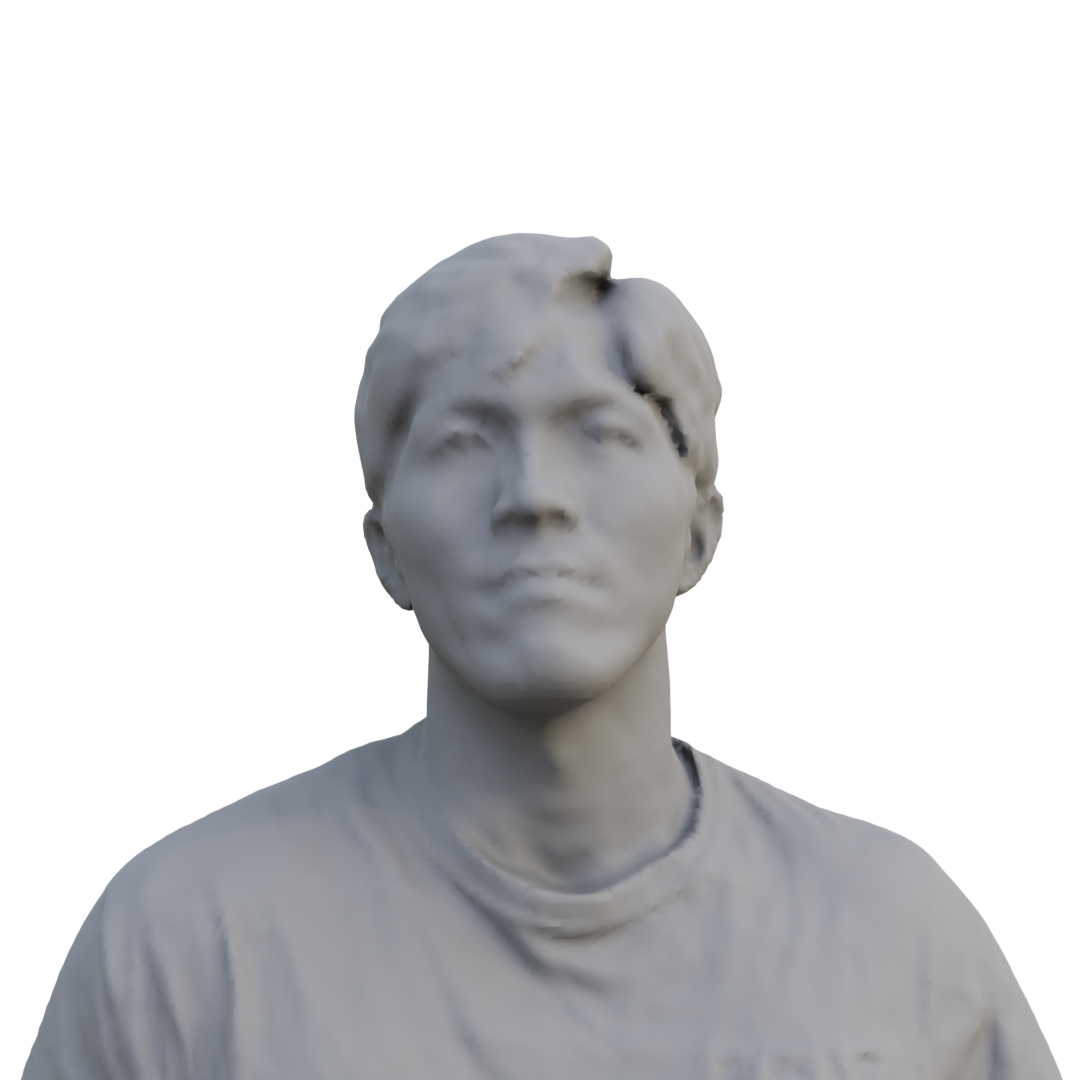} &
    \includegraphics[width=0.12\linewidth]{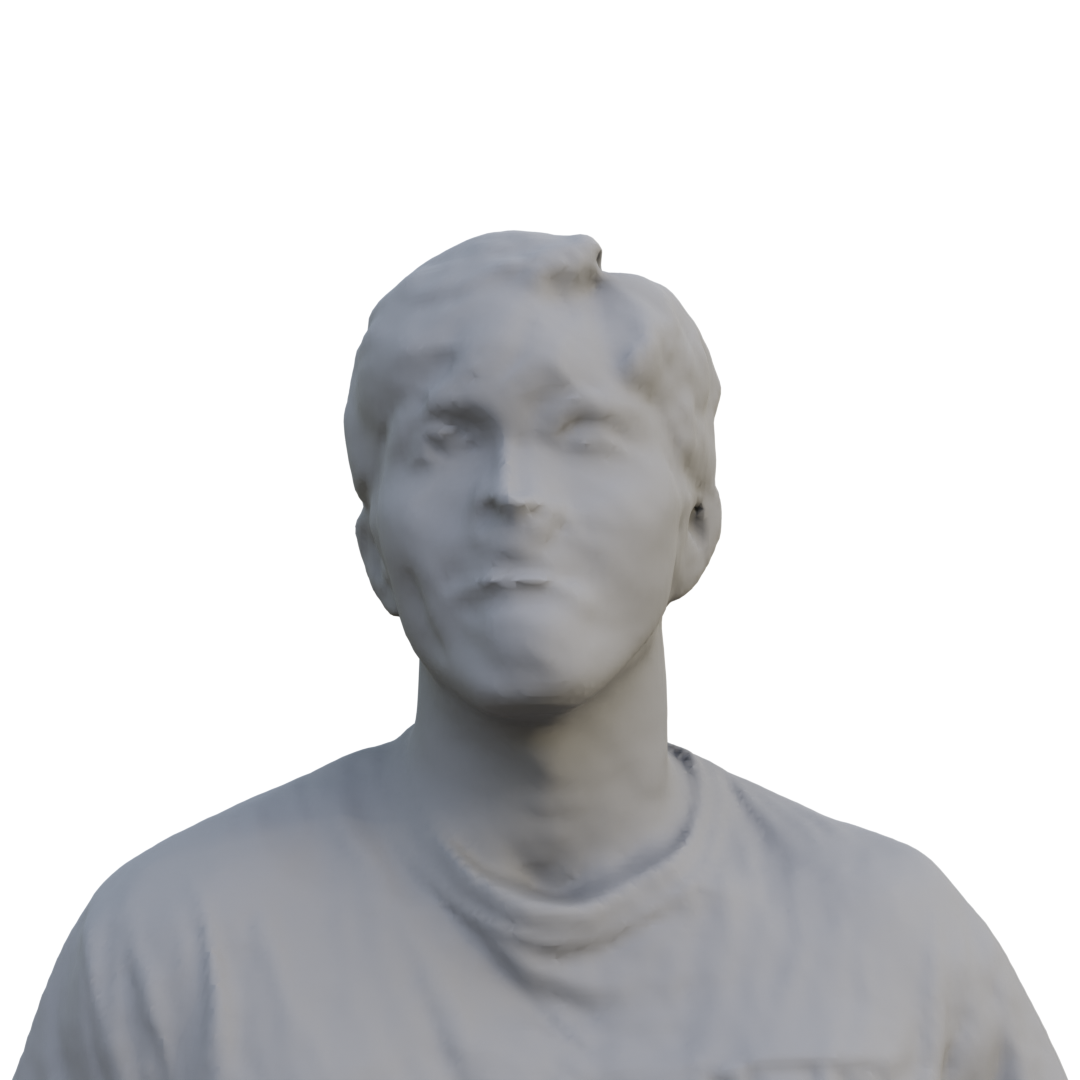} &
    \includegraphics[width=0.12\linewidth]{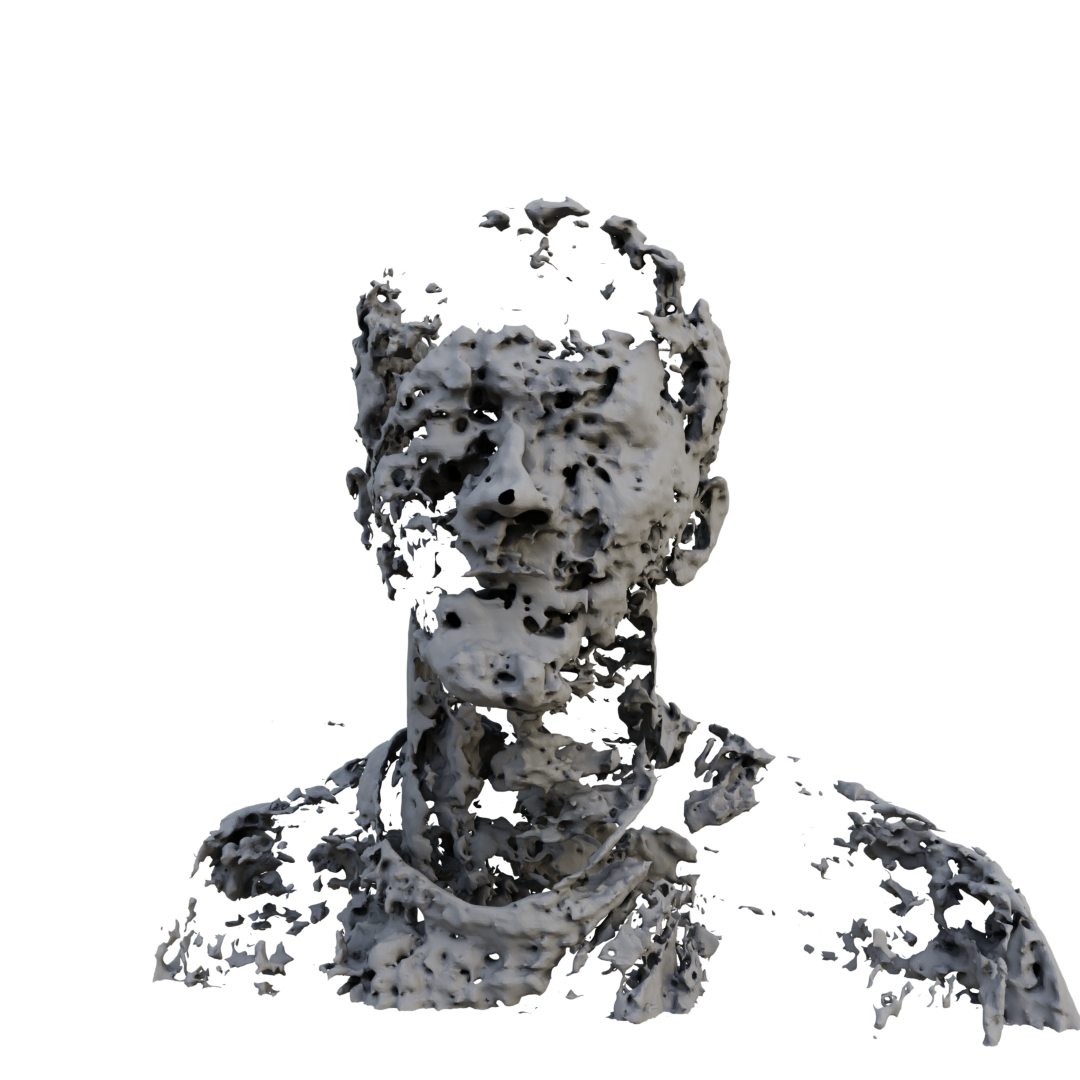} &
    \includegraphics[width=0.12\linewidth]{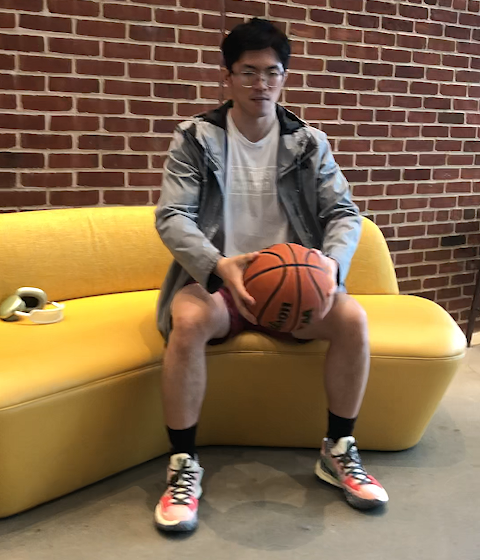}  &
    \includegraphics[width=0.12\linewidth]{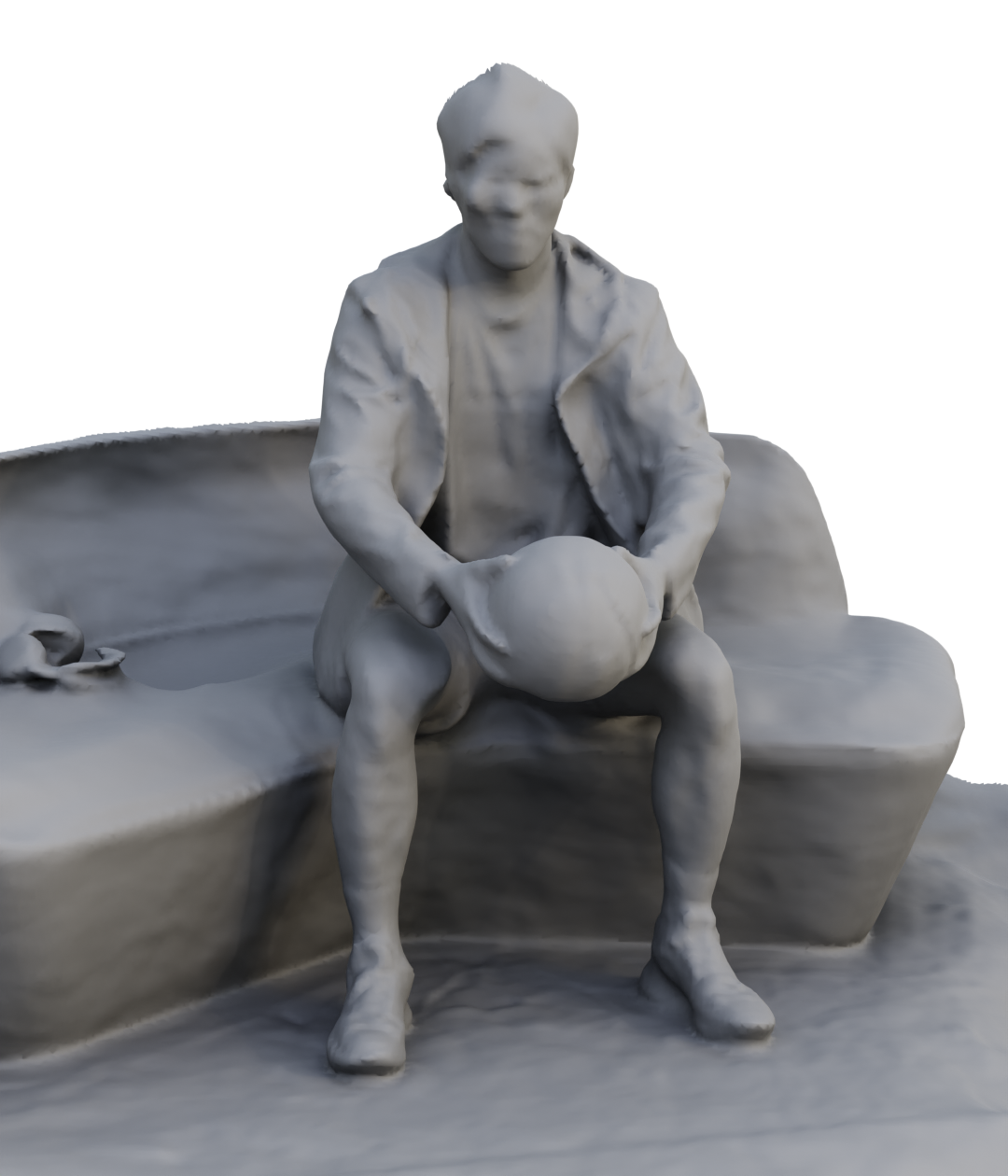}  &
    \includegraphics[width=0.12\linewidth]{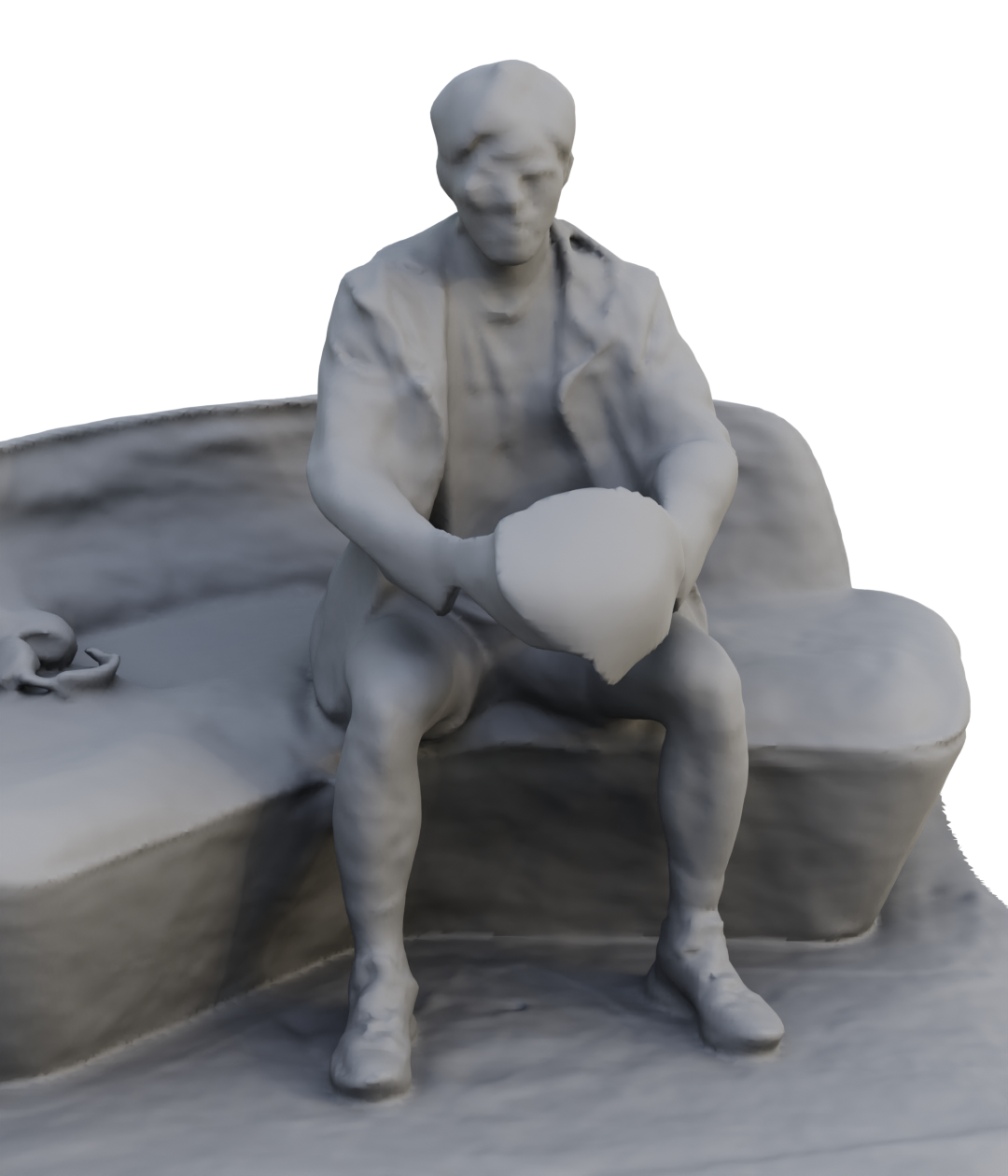} &
    \includegraphics[width=0.12\linewidth]{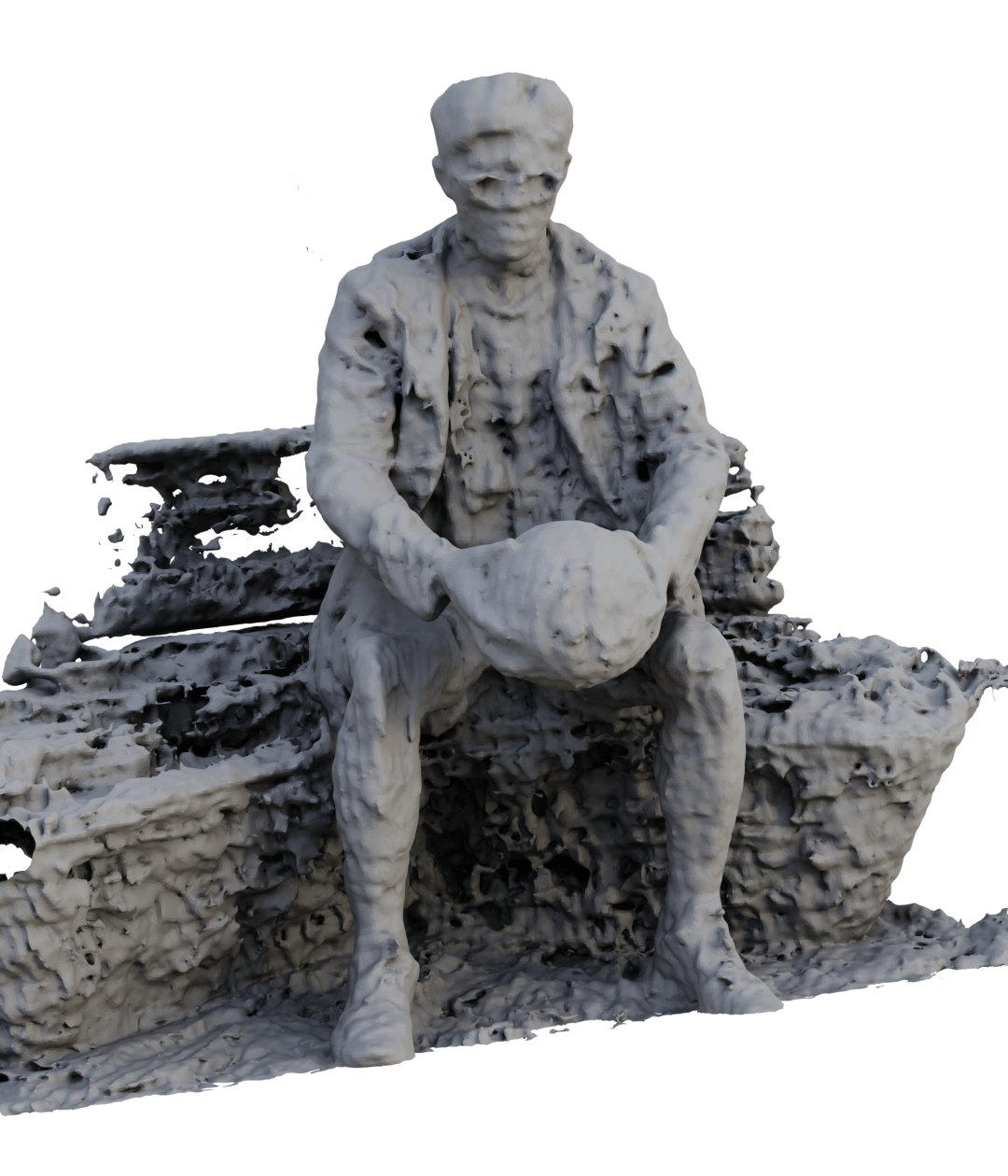} 
    \\
  \scriptsize (a) Input  & \scriptsize (b) Ours &  \scriptsize (c)  NeuS  & \scriptsize (d) D-NeRF  & \scriptsize (e) Input  & \scriptsize (f) Ours &  \scriptsize (g) NeuS   & \scriptsize (h) D-NeRF   
\end{tabular}
\vspace{-0.5em}
\caption{{\bf Surface reconstruction from dynamic videos}. Our method achieves higher-quality, more detailed reconstruction (e.g. on muscles, faces, and books) than other approaches.} 
\vspace{-4mm}
\label{fig:geometry}
\end{figure}

We visualize mesh reconstruction quality of dynamic scenes in Figure~\ref{fig:geometry}. Table~\ref{tab:recon} shows quantitative results. Our method frequently outperforms competitive approaches. Additional geometry reconstruction comparisons can be found in Appendix~\ref{app:video}.


\vspace{-0.5em}
\subsection{Novel View Synthesis}
\label{sec:novel}
\vspace{-0.5em}

\begin{figure}
\centering
\begin{tabular}{@{}c@{\hspace{0.3mm}}c@{\hspace{0.3mm}}c@{\hspace{0.3mm}}c@{\hspace{0.3mm}}c@{\hspace{0.3mm}}c@{\hspace{0.3mm}}@{}}
    \includegraphics[width=0.16\linewidth]{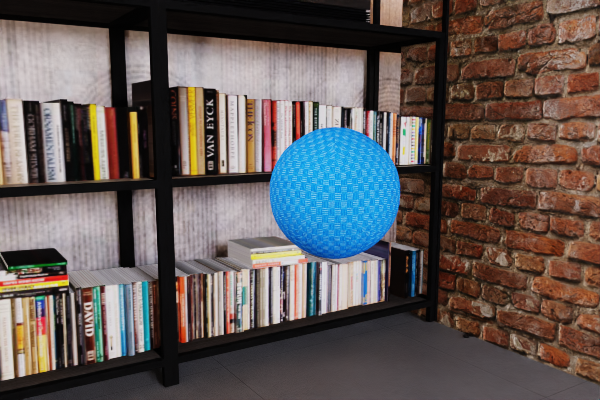} &
    \includegraphics[width=0.16\linewidth]{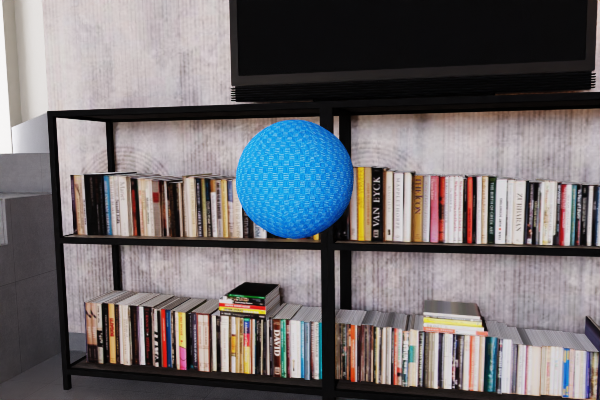} &
    \includegraphics[width=0.16\linewidth]{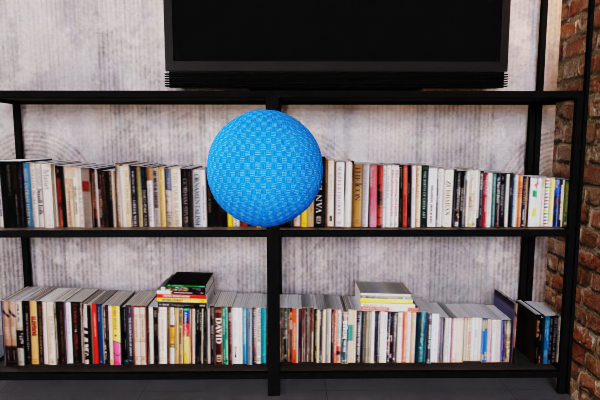} &
    \includegraphics[width=0.16\linewidth]{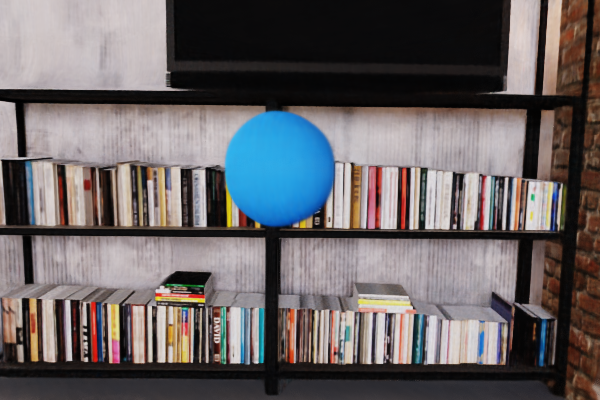} &
    \includegraphics[width=0.16\linewidth]{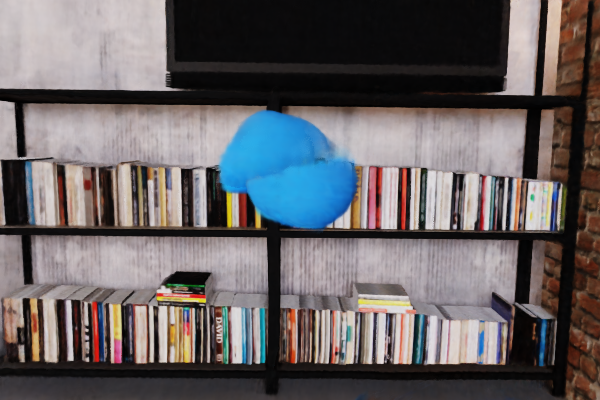} &
    \includegraphics[width=0.16\linewidth]{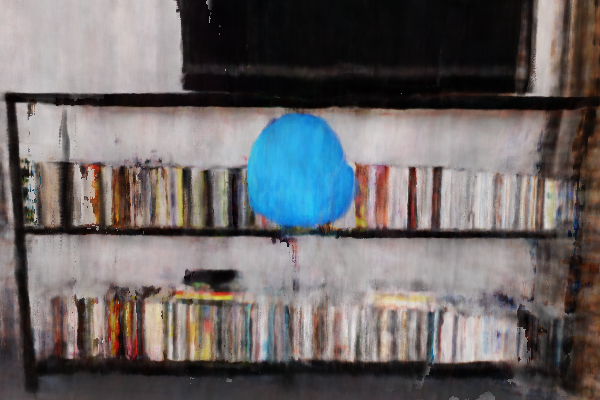} \\
    \includegraphics[width=0.16\linewidth]{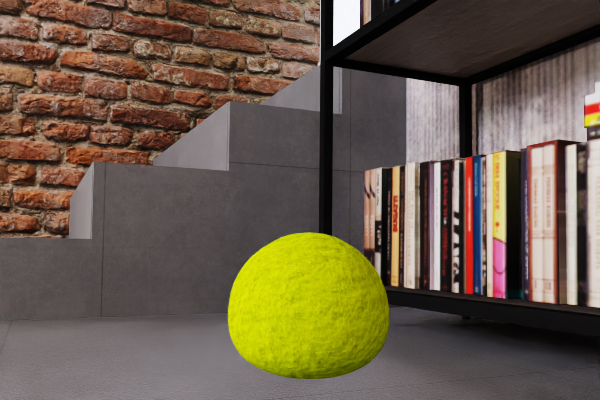} &
    \includegraphics[width=0.16\linewidth]{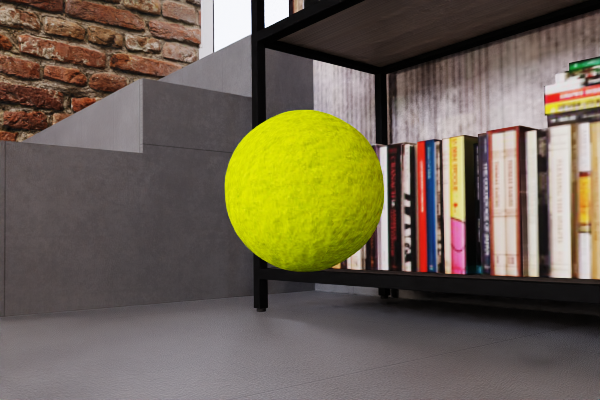} &
    \includegraphics[width=0.16\linewidth]{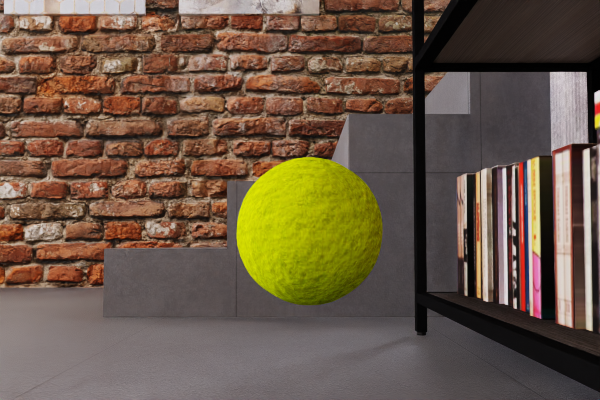} &
    \includegraphics[width=0.16\linewidth]{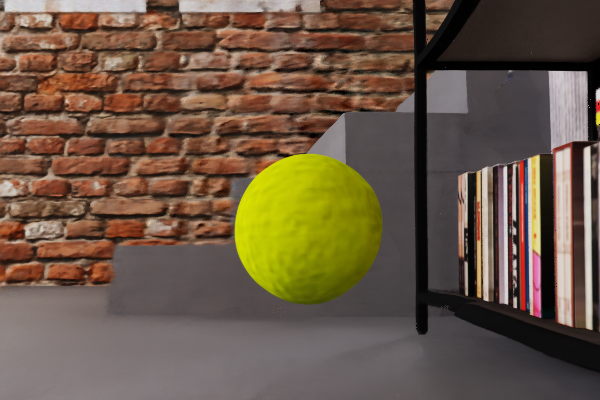} &
    \includegraphics[width=0.16\linewidth]{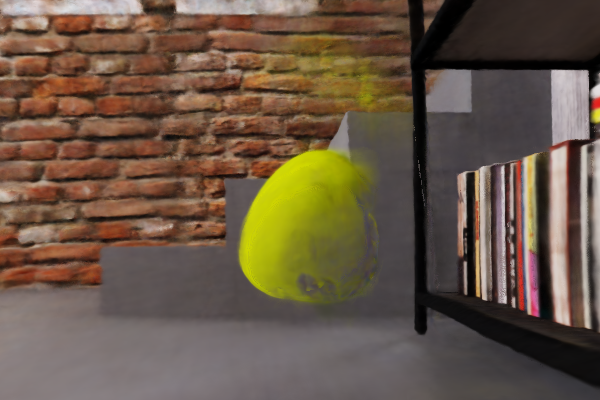} &
    \includegraphics[width=0.16\linewidth]{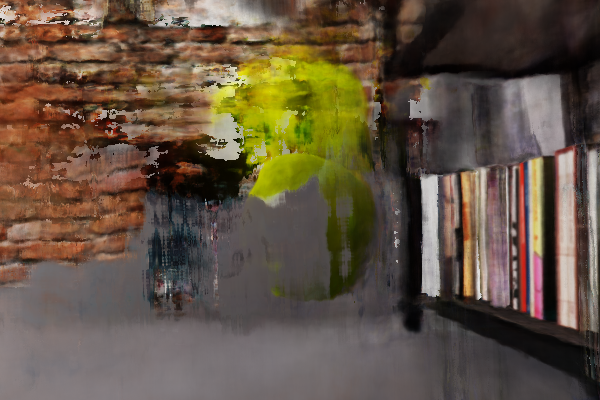}  \\
   \scriptsize (a) Input 1  & \scriptsize (b) Input 2 &  \scriptsize (c) Novel view on 2   & \scriptsize (d) Ours  & \scriptsize (e) NRNeRF & \scriptsize (f) D-NeRF 
\end{tabular}
\vspace{-0.5em}
\caption{{\bf Qualitative evaluation of geometry reconstruction and novel view synthesis}. (a, b) Frames from the training videos. (c) Ground truth evaluation view. Consistent with numerical results in Table~\ref{tab:novel}, our method generates images from novel views with higher quality than previous methods. }
\vspace{0em}
\label{fig:novel}
\end{figure}

We simulate four cases of deformable objects moving in a room. We render each scene with two different camera trajectories, so that one path may be used for training, while the other is used as a ground truth to evaluate novel view synthesis. Quantitative results in Table~\ref{tab:novel} show that our method achieves the best synthesis quality in all metrics. Datasets and results are shown in Figure~\ref{fig:novel}. (a) and (b) are frames from the training videos. (c) is taken at the same time step as (b), but it is a novel view from the test path. Notice that our method successfully synthesizes the dynamic scene from a novel view, while other methods struggle to reconstruct geometry and texture accurately.

\begin{table*}
\newcolumntype{Z}{S[table-format=2.3,table-auto-round]}
\centering
\setlength{\tabcolsep}{3mm}
\ra{1.05}
\small
\resizebox{1\linewidth}{!}{
\begin{tabular}{@{}lcccccccccccc@{}}
  \toprule
  \multirow{2}[3]{*}{Method}  & \multicolumn{3}{c}{Tennis Ball}  &  \multicolumn{3}{c}{Basketball }   &  \multicolumn{3}{c}{Blue Ball}  &  \multicolumn{3}{c}{Blue Cube}  \\
  \cmidrule(l{3mm}r{3mm}){2-4} \cmidrule(l{3mm}r{3mm}){5-7} \cmidrule(l{3mm}r{3mm}){8-10} \cmidrule(l{3mm}r{3mm}){11-13}
  & LPIPS$\downarrow$ & SSIM$\uparrow$ & PSNR$\uparrow$& LPIPS$\downarrow$ & SSIM$\uparrow$ & PSNR$\uparrow$& LPIPS$\downarrow$ & SSIM$\uparrow$ & PSNR$\uparrow$& LPIPS$\downarrow$ & SSIM$\uparrow$ & PSNR$\uparrow$   \\
  \midrule
  NRNeRF   & 0.455  & 0.566  & 30.286&  0.473 & 0.550 & 29.684 &  0.409 & 0.479 & 30.152 &  0.620 & 0.398 & 29.044\\
  D-NeRF   & 0.596  & 0.392  & 29.170&  0.565 & 0.425 & 29.073 & 0.596  & 0.317 & 28.937 &  0.775 & 0.229 & 27.855\\
  Ours   & \textbf{0.336}  &\textbf{ 0.632}  & \textbf{31.103}&  \textbf{0.368} & \textbf{0.625} & \textbf{30.568} & \textbf{ 0.334} & \textbf{0.513} & \textbf{30.407} &  \textbf{0.589} & \textbf{0.449} & \textbf{29.997}\\
   \bottomrule
\end{tabular}
}
\vspace{-0.5em}
\caption{{\bf Quantitative evaluation of novel view synthesis}. In this experiment, we evaluate a novel camera trajectory different from the training data. Even in this highly under-constrained task, our method benefits from the consistency of the geometry and significantly outperforms S.O.T.A. using all metrics. }
\label{tab:novel}
\end{table*}

\begin{figure}
\centering
\resizebox{\linewidth}{!}{
\begin{tabular}[c]{@{}c@{\hspace{0.2mm}}c@{\hspace{0.2mm}}c@{\hspace{0.2mm}}c@{\hspace{0.2mm}}c@{\hspace{0.2mm}}c@{\hspace{0.2mm}}@{}}
    \includegraphics[width=0.20\linewidth]{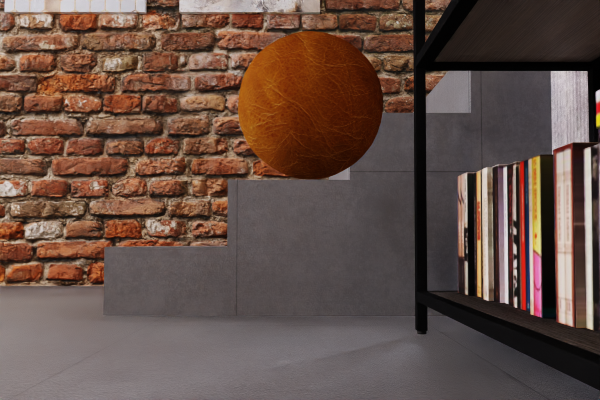} &
    \includegraphics[width=0.20\linewidth]{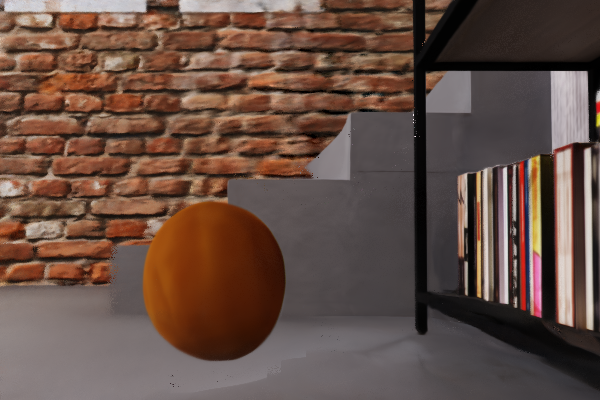} &
    \includegraphics[width=0.20\linewidth]{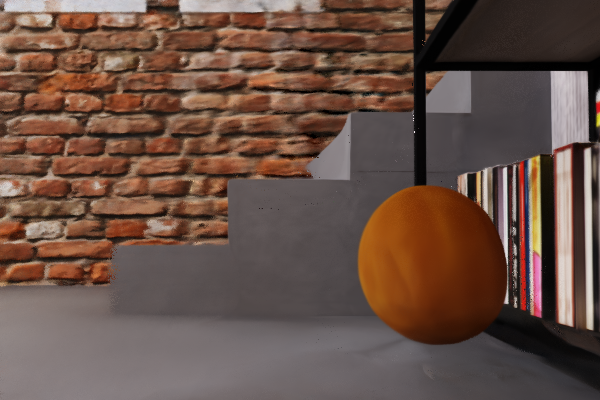} &
    \includegraphics[width=0.20\linewidth]{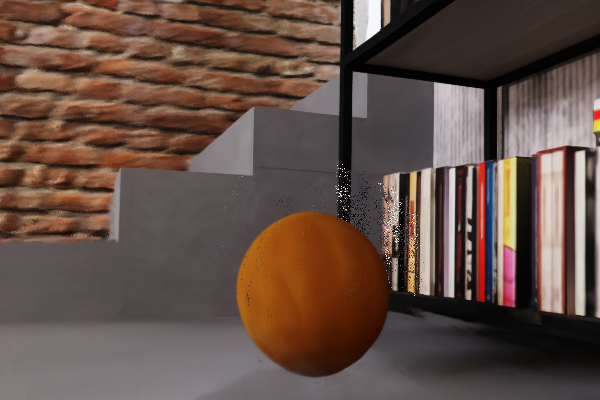}&
    \includegraphics[width=0.20\linewidth]{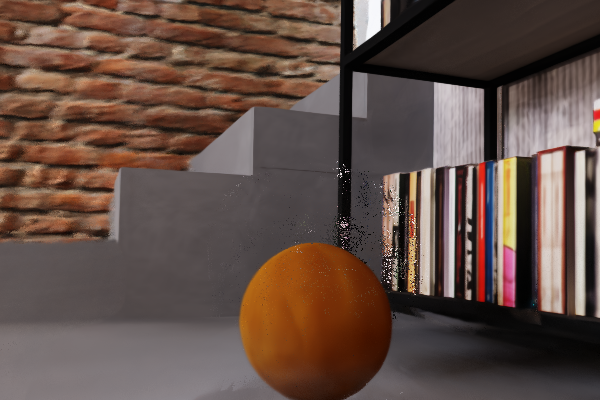}&
    \includegraphics[width=0.20\linewidth]{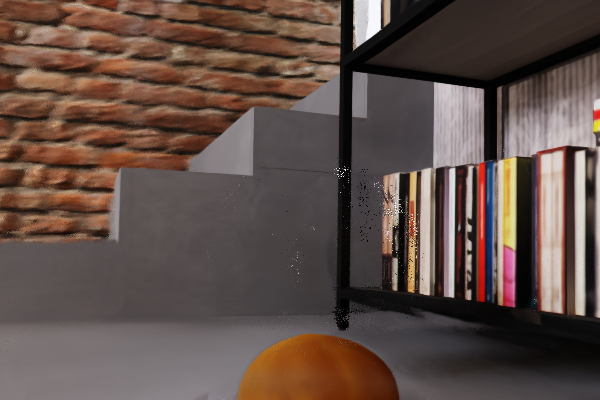}\\
   \small (a) Original Image   & \small (b) Left velocity   &\small (c) Right velocity   & \small (d) Harder material   & \small (e) Moderate material   & \small (f) Softer material    \\ 
\end{tabular}
}
\vspace{-0.5em}
\caption{{\bf Parameter estimation and editing by differentiable physics}. NeuPhysics can estimate the acceleration and Young's Modulus, generate new videos with different initial velocities and material parameters.}
\label{fig:estimate}
\vspace{-1em}
\end{figure}


\subsection{Parameter Estimation}
\label{sec:estimate}
In these experiments, we first use our network to reconstruct the dynamic scene, then apply the embedded differentiable simulator to estimate physical parameters. Finally, we edit the physical parameters to generate novel video frames. More visual results can be found in the supplementary video. Figure~\ref{fig:estimate} shows the initial frame of a bouncing ball sequence, where the ball is released from a height and collides with the ground. The ball is modeled by co-rotated material. Using DiffPD~\cite{du2021diffpd}, we estimate both Young's modulus and vertical acceleration in two separate optimization tasks. The physics loss is minimized for 100 epochs (as described in Section~\ref{sec:simulation}) using Adam~\cite{kingma2014adam} optimizer with a learning rate of 0.01. We initialize Young's modulus to $2\times 10^5 Pa$ and acceleration to $0m/s^{2}$, and by the end of optimization, have estimated Young's modulus at $2.96\times 10^6 Pa$ and acceleration at $3.78m/s^2$. 
Having identified accurate physics parameters, we may choose to modify some of them from their true values, and generate videos that contain modified physical properties from the original on which they are based. For example, the original ball has no horizontal speed, so we add $0.3m/s$ left and right initial velocities in Figure~\ref{fig:estimate} (b) and (c), respectively. We also decrease the Young's modulus in Figure~\ref{fig:estimate} (d), (e), and (f) from $2\times 10^7 Pa$ to $2\times 10^5 Pa$, where the deformation is getting more significant.

\begin{figure}
\centering
\begin{tabular}{@{}c@{\hspace{0.3mm}}c@{\hspace{0.3mm}}c@{\hspace{0.3mm}}c@{\hspace{0.3mm}}c@{\hspace{0.3mm}}@{}}
    \includegraphics[width=0.19\linewidth]{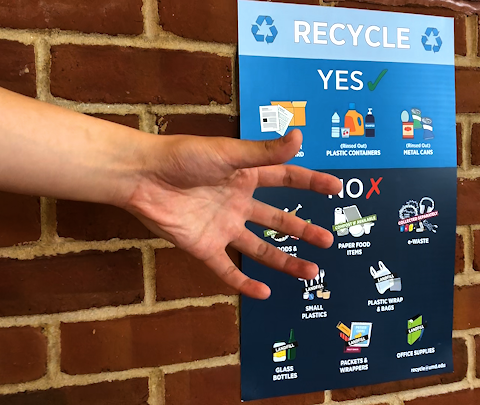} &
    \includegraphics[width=0.19\linewidth]{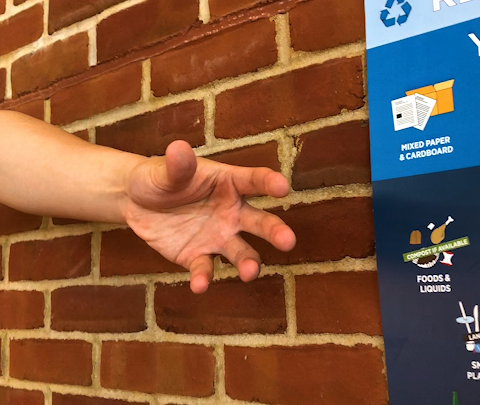} &
    \includegraphics[width=0.19\linewidth]{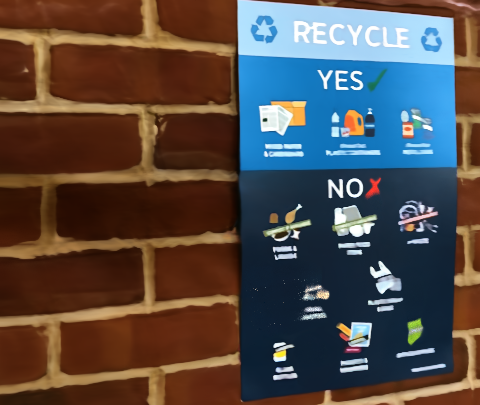} &
    \includegraphics[width=0.19\linewidth]{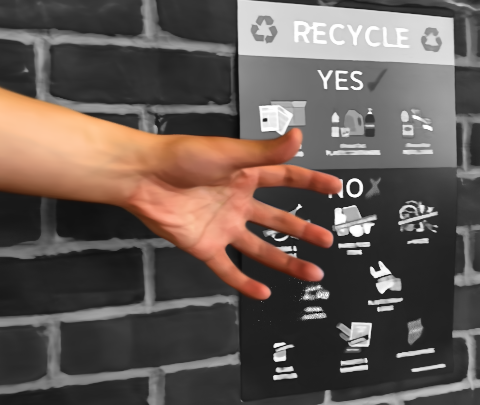} &
    \includegraphics[width=0.19\linewidth]{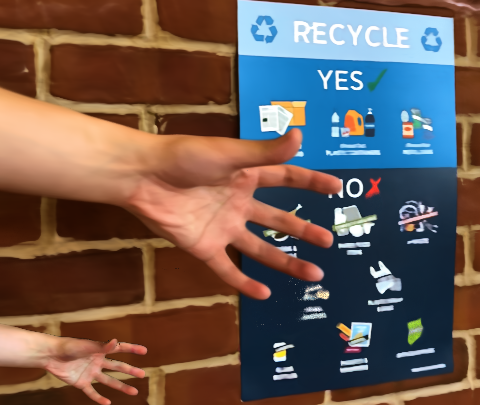} \\
    \includegraphics[width=0.19\linewidth]{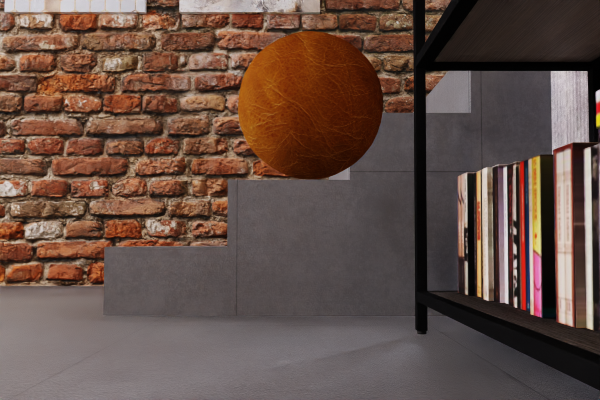} &
    \includegraphics[width=0.19\linewidth]{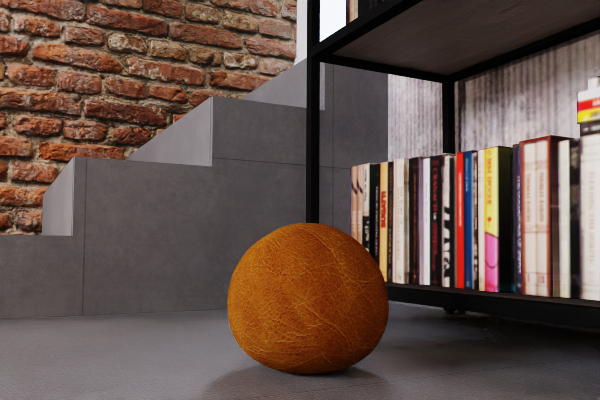} &
    \includegraphics[width=0.19\linewidth]{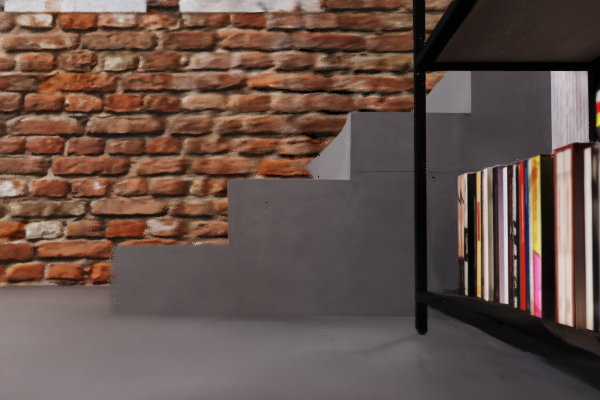} &
    \includegraphics[width=0.19\linewidth]{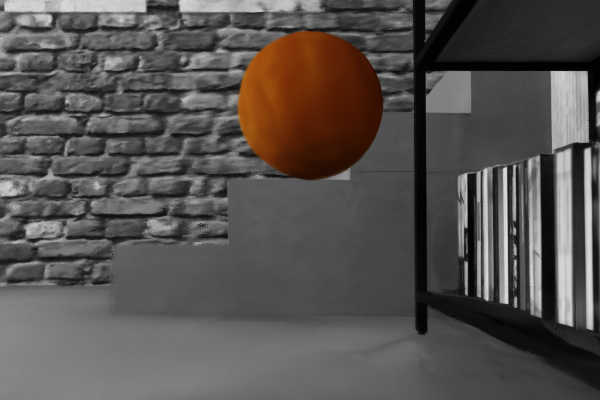} &
    \includegraphics[width=0.19\linewidth]{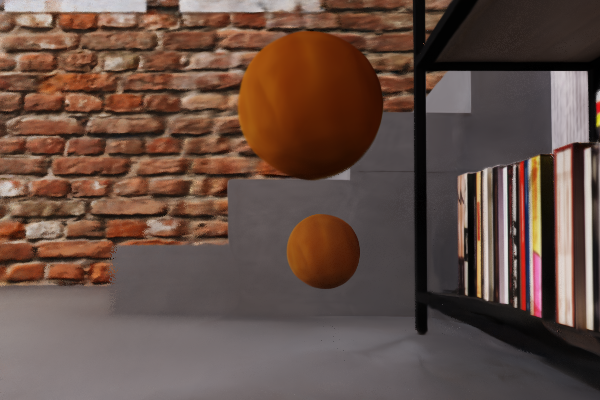}  \\
   \small (a) View 1   & \small (b) View 2 & \small (c) Delete  & \small (d) Appearance & \small (e) Duplicate 
\end{tabular}
\vspace{-0.5em}
\caption{{\bf Geometry and appearance editing in videos}. Our method can delete, add, move, or change the appearance. Editing in the 3D space can provide users with realistic and clean results. }
\vspace{-4mm}
\label{fig:edit_qual}
\end{figure}

\subsection{Editing}
\label{sec:edit}
In this section, we demonstrate how our hybrid representation provides a flexible interface for users to directly modify the geometry on the 3D hexahedral mesh and volume render the results.

\mypara{Region of Interest.} First, we select a region of interest that we wish to edit. Our system naturally provides the scene rigidity mask and SDF values, so a foreground area $\mathcal{A}_i$ at the $i$-th frame can be computed as Equation~\ref{eq:foreground}. If users require a more precisely defined region of interest, a custom bounding box or even more complex hexahedra bounding volume $\mathcal{A}'$ can be provided. A hexahedra mesh $\mm$ can be reconstructed from the final region of interest, $\mathcal{A}' \cap \mathcal{A}_i$,. 

\mypara{Delete.} Figure~\ref{fig:edit_qual} (a) and (b) are two distinct time steps from a given input video. In one, a hand flexes and deforms, and in the second, an elastic ball bounces. First, we delete the foreground from View 1 by simply setting the SDF value to a positive number (e.g. 1.0) for the query points inside the hexahedral mesh $\mm$. Since we assume a bell-shaped density distribution centered at 0~\cite{wang2021neus}, an area with large SDF value is less likely to be sampled, and therefore appears transparent. After removing the foreground from View 1, even though some parts of the background were occluded in the original, the other views of the scene provide sufficient textural/structural information to complete the missing background. Since editing takes place entirely in 3D space, the resulting 2D images feel sharp and physically grounded, as shown in (c).

\mypara{Appearance.} We can also modify the output of the color network to change the appearance of a specific region. For example, we might like to emphasize the moving foreground of a video by desaturating the background. This is easily done by changing the original color $[r,g,b]=\cc_{\theta_c}(\pp)$ to be $[\frac{r+g+b}{3}, \frac{r+g+b}{3}, \frac{r+g+b}{3}]$ for all the query points outside $\mm$. Results are shown in Figure~\ref{fig:edit_qual} (d).

\mypara{Duplicate.} Translation and scaling are basic operations in graphic design software. With our pipeline, users are able to copy, move, and scale the hexahedral mesh $\mm$, creating a new version $\mm'$. During volume rendering, the SDF and color values of the points $\pp'$ inside $\mm'$ are set equal to their counterparts $\pp$ in $\mm$. In Figure~\ref{fig:edit_qual} (d), we duplicate an existing object and modify its size and position. 

\begin{wrapfigure}{r}{0.4\linewidth}
\centering
\resizebox{1\linewidth}{!}{
\begin{tabular}{l|cc}
	\toprule
	 & Rendering Module & Physics Module  \\
	\hline
	Epoch  & 300,000 & 20   \\
	Time per epoch  & $\sim 0.22$ s & $\sim 70$ s  \\
	Total time& $\sim 18$ h  & $\sim 0.5$ h \\
	\bottomrule
\end{tabular}
}
\resizebox{1\linewidth}{!}{
\begin{tabular}{@{}c@{}}
    \includegraphics[width=\linewidth]{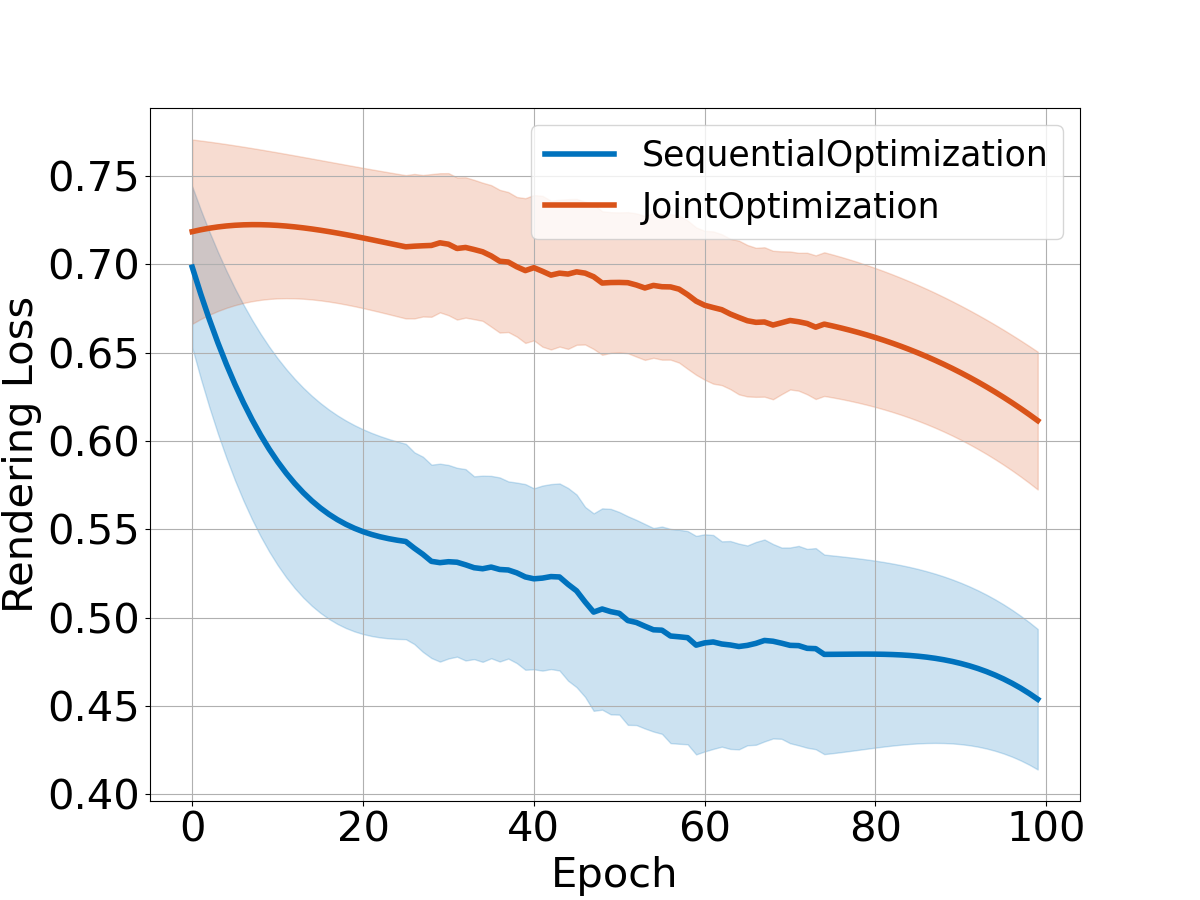} 
\end{tabular}
}
\vspace{-1em}
\caption{Joint training can be time-consuming and converges slowly. }
\vspace{-2mm}
\label{fig:joint}
\end{wrapfigure}

\subsection{Ablation Study}
\label{sec:ablation}
In this section, we conduct several ablation studies using the bouncing ball example to justify our network design choices and training strategies.

\mypara{Sequential Vs. Joint Optimization.} Our method learns the scene sequentially by first optimizing the rendering networks, followed by the physics parameters. A seemingly more concise strategy might be to jointly train all parameters from scratch. Unfortunately, the joint training strategy is impractical for two reasons: (1) the table in Figure~\ref{fig:joint} indicates that the physics module converges rapidly thanks to the effectiveness of the differentiable simulator, but running the forward and backward simulation is costly ($\sim 70$ s) compared to querying the rendering networks ($\sim 0.22$ s) -- training the rendering module and physics module together would require many more iterations (300,000 epochs), with each taking much longer ($\sim 70$sec per epoch), making the training infeasible, and (2) before the rendering module converges and provides satisfactory geometry estimation, the geometry makes little sense for simulation, so gradients flowing from the $\mathcal{L}_{physics}$ do not help the optimization. The smoothed plot in Figure~\ref{fig:joint} indicates that joint training slows down the rendering optimization early on. Sequential optimization is therefore more practical. 

We conducted two additional tests to study the relationship between the rendering and physics modules. First, during the sequential optimization, we also update the rendering networks. This harms the rendering quality (increases loss from 0.03 to 0.10) and does not help physics parameter estimation. Second, we attempt to optimize the rendered image from physics simulation to match the ground truth image. We found this process to be inefficient (2 minutes per image) and observed no significant benefits.

\begin{figure}
\centering
\resizebox{0.6\linewidth}{!}{
\begin{tabular}[c]{@{}c@{\hspace{0.3mm}}c@{\hspace{0.3mm}}c@{}}
    \includegraphics[width=0.32\linewidth]{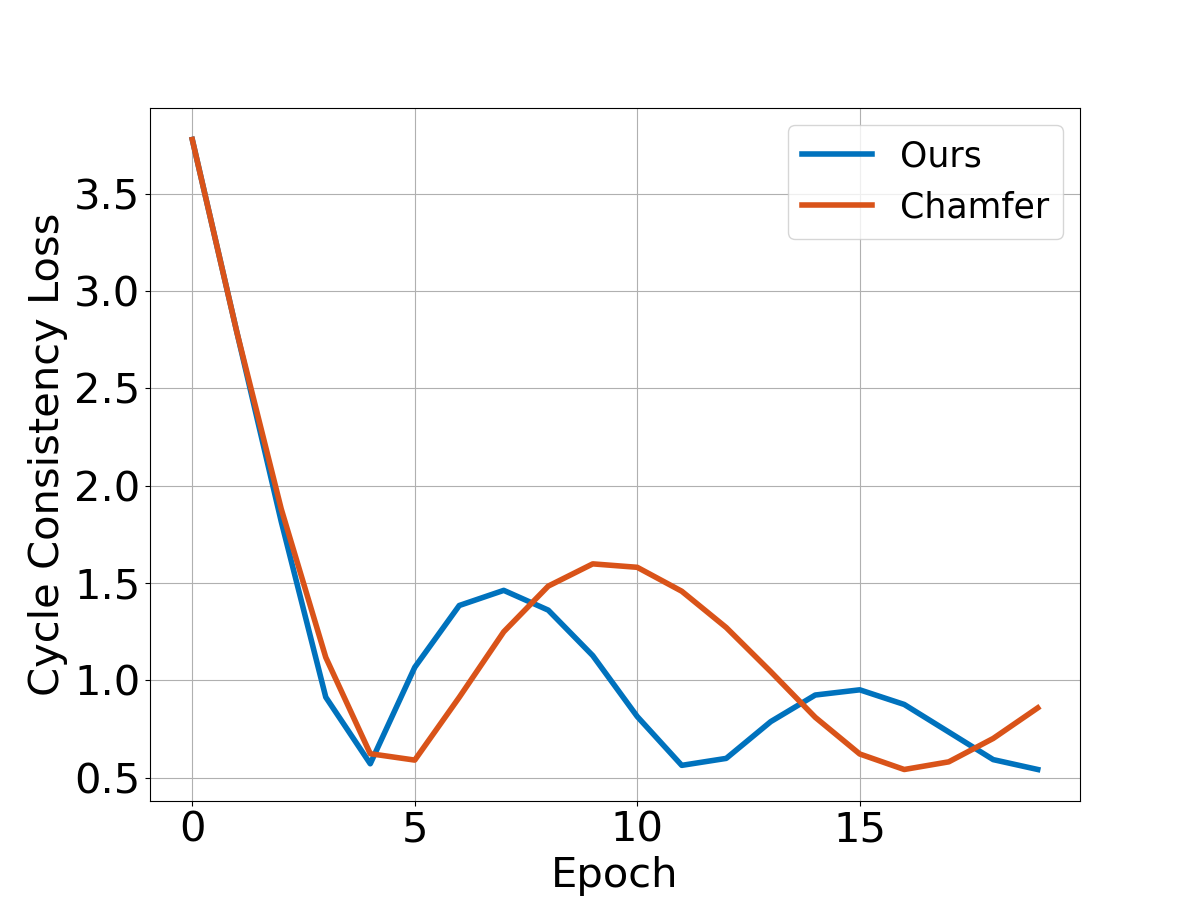} &
    \includegraphics[width=0.32\linewidth]{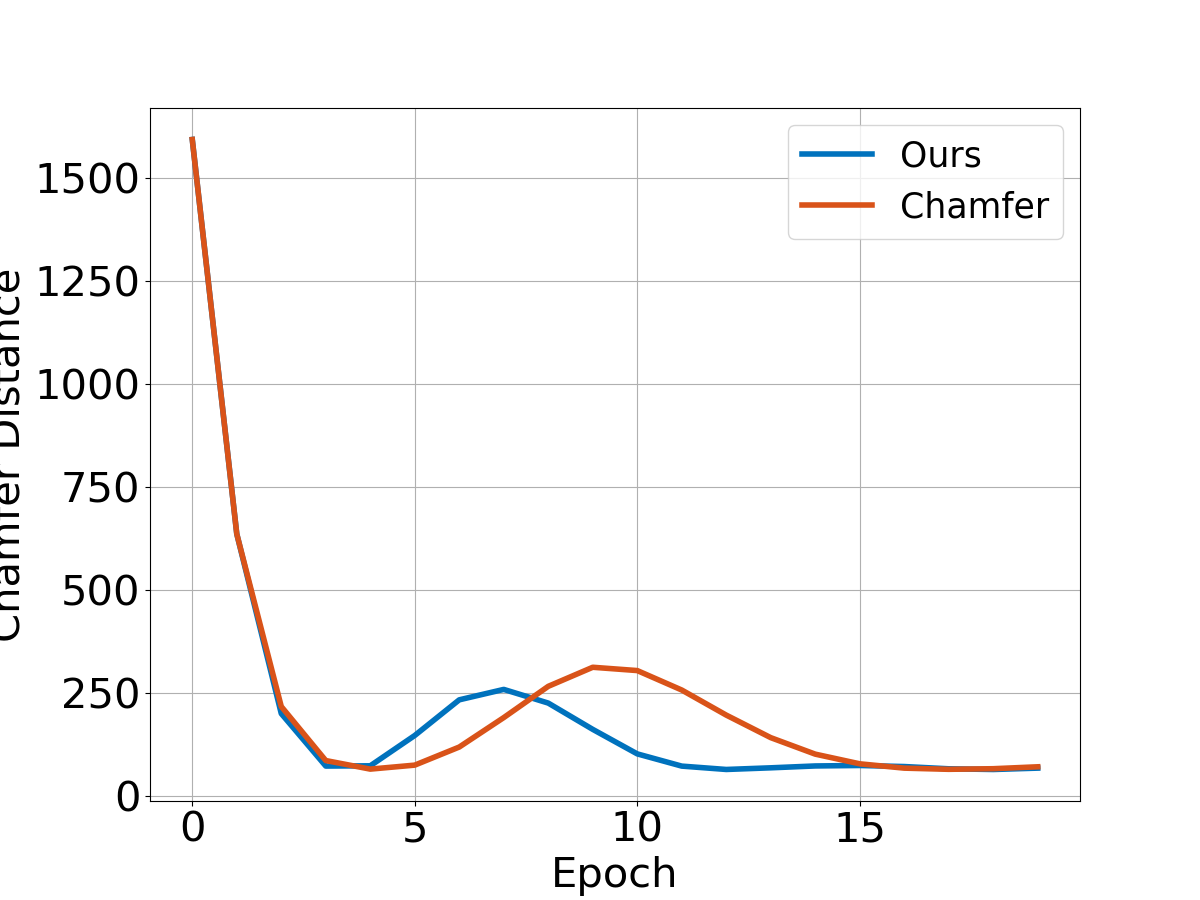} &
    \includegraphics[width=0.32\linewidth]{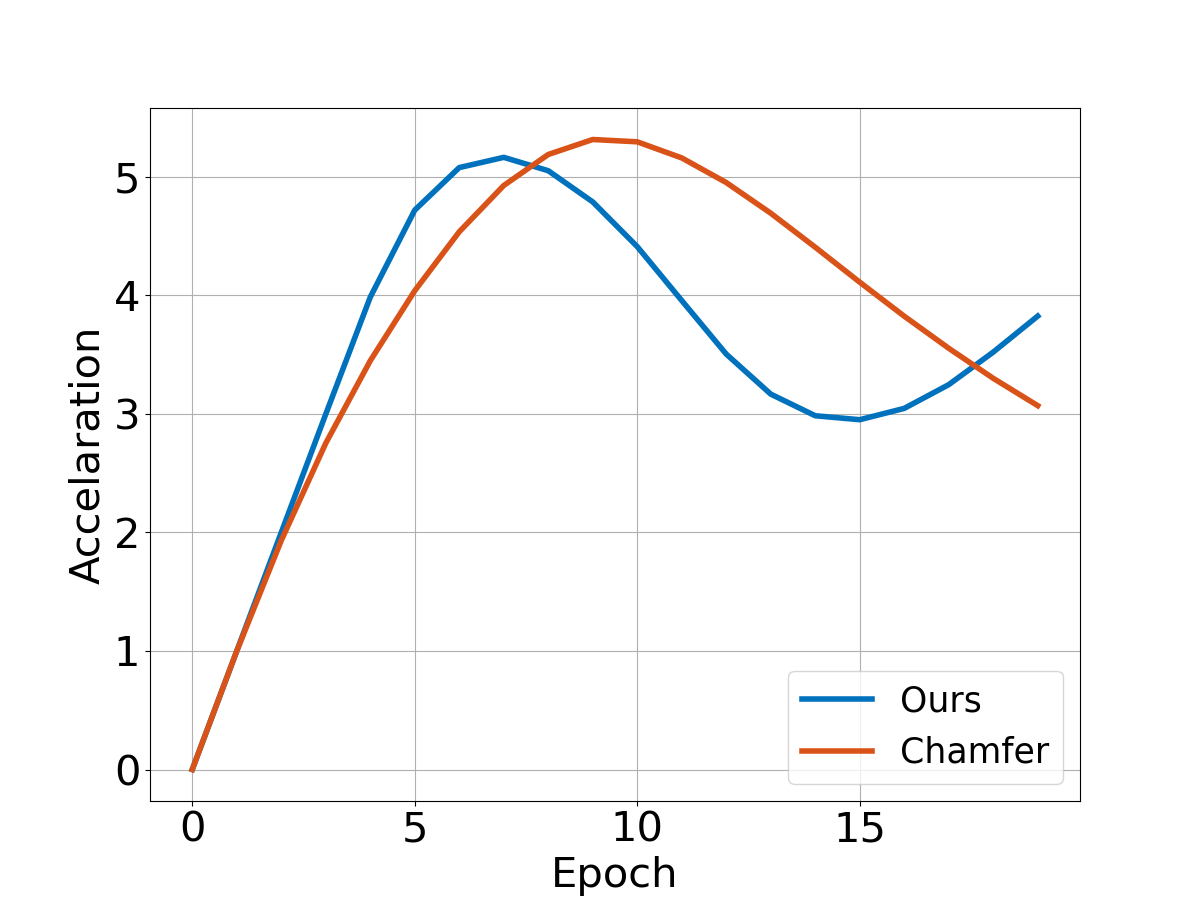}\\
   \small (a) Cycle Consistency Loss  & \small (d) Chamfer Distance Loss  & \small (f) Acceleration  
\end{tabular}
}
\resizebox{0.35\linewidth}{!}{
\begin{tabular}[c]{l|cc}
	\toprule
	Time & Ours & Reconstruct+Chamfer  \\
	\hline
	Reconstruction  & $\sim 2$ s &  $\sim 50$ s  \\
	Simulation  & $\sim 65$ s & $\sim 65$ s  \\
	Total & $\sim 67$ s  & $\sim 115$ s \\
	\bottomrule
\end{tabular}
}
\caption{\nps{{\bf Compare with optimizing Chamfer Distance.} An alternative to our proposed cycle-consistency loss is to use Chamfer Distance for the physics module. The two metrics have similar effects but our method can save much reconstruction time.}}
\label{fig:physics}
\vspace{-1.5em}
\end{figure}

\nps{
\mypara{Physics Module.} Within our sequential optimization strategy, there are still multiple options for physical parameter estimation. In our pipeline, we reconstruct the volume mesh $\mm_1$ of the first frame, then predict the following mesh sequence $\mm'_i = sim(\theta_{physics}, \mm_1, 1 \rightarrow i)$ with the differentiable simulator. The physics loss for each frame can be computed in a cycle-consistency manner as in Equation~\ref{eq:cycle}. Our proposed cycle-consistency loss naturally guarantees a point-wise correspondence for each vertex, and does not require mesh reconstruction for every time step. For each epoch, we only reconstruct the volume mesh for the first frame, which reduces computation and makes the loss differentiable w.r.t. the rendering networks.\\\\
A more direct physics loss could rely on reconstructing the geometry of each frame and computing the Chamfer distance between the simulated and reconstructed meshes. We compare the two strategies in Figure~\ref{fig:physics}. `Ours' is optimized with our proposed cycle-consistency loss $\mathcal{L}_{physics}$ and `Chamfer' is optimized with the Chamfer distance. (c) shows the value of \textit{acceleration}, the optimization variable. (a) and (b) indicate that the losses have comparable performance in both metrics. However, `Chamfer' takes significantly longer since every frame must be reconstructed. Our proposed cycle-consistency loss runs faster and maintains differentiability between the rendering and physics simulation module. Conceptually, decoupling the NeRF and differentiable physics has another drawback: a mesh reconstructed purely from SDF cannot capture the interior \textit{stress and deformation}. Our proposed cycle-consistency loss addresses this concern, as it takes advantage of the learned motion field, so it is capable of finding the point-wise correspondence and describing the deformation.
} 

\begin{wrapfigure}{r}{0.5\linewidth}
\centering
\resizebox{1\linewidth}{!}{
\begin{tabular}{@{}c@{\hspace{0.3mm}}c@{\hspace{0.3mm}}c@{}}
    \includegraphics[width=0.32\linewidth]{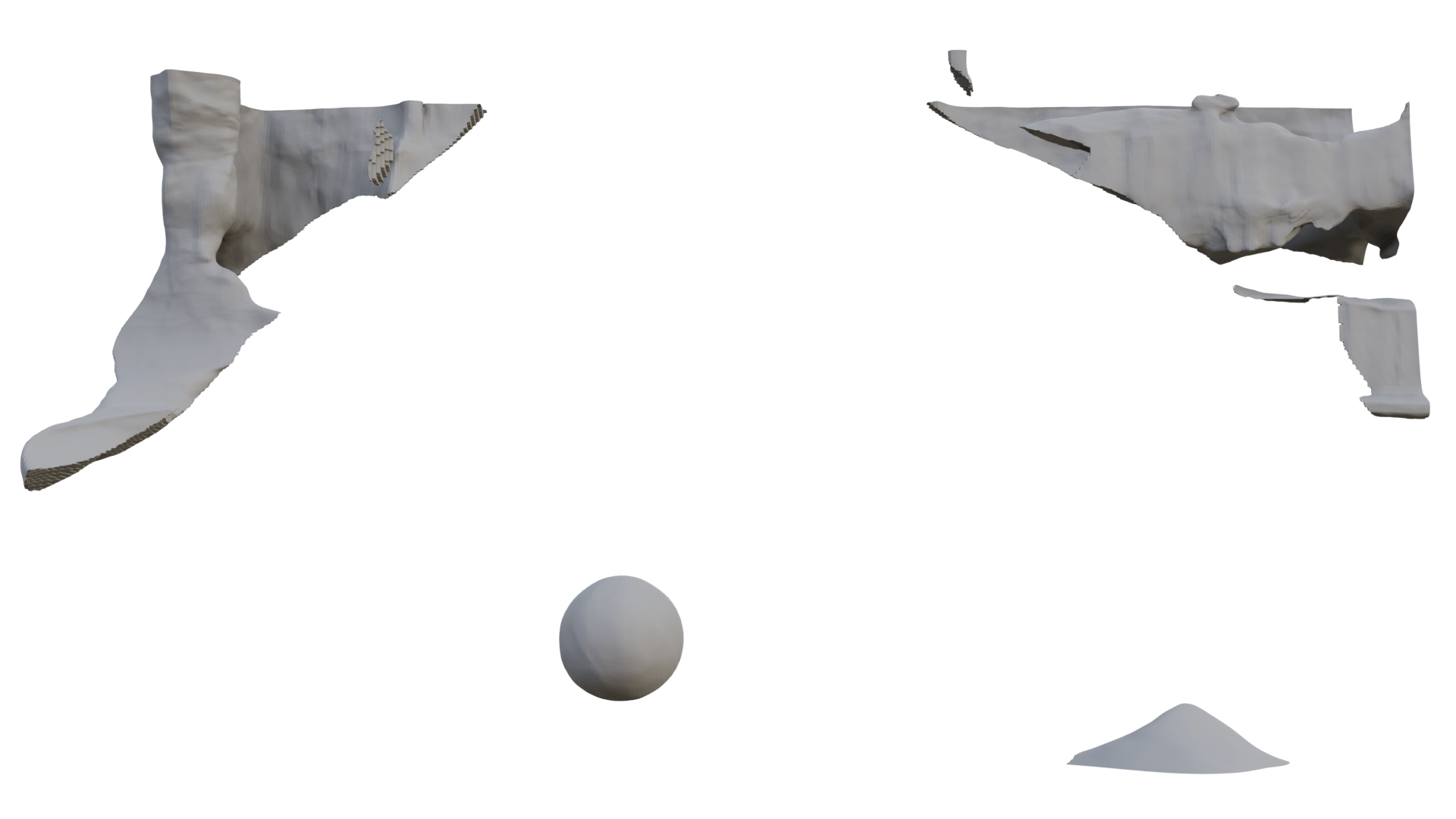} &
    \includegraphics[width=0.32\linewidth]{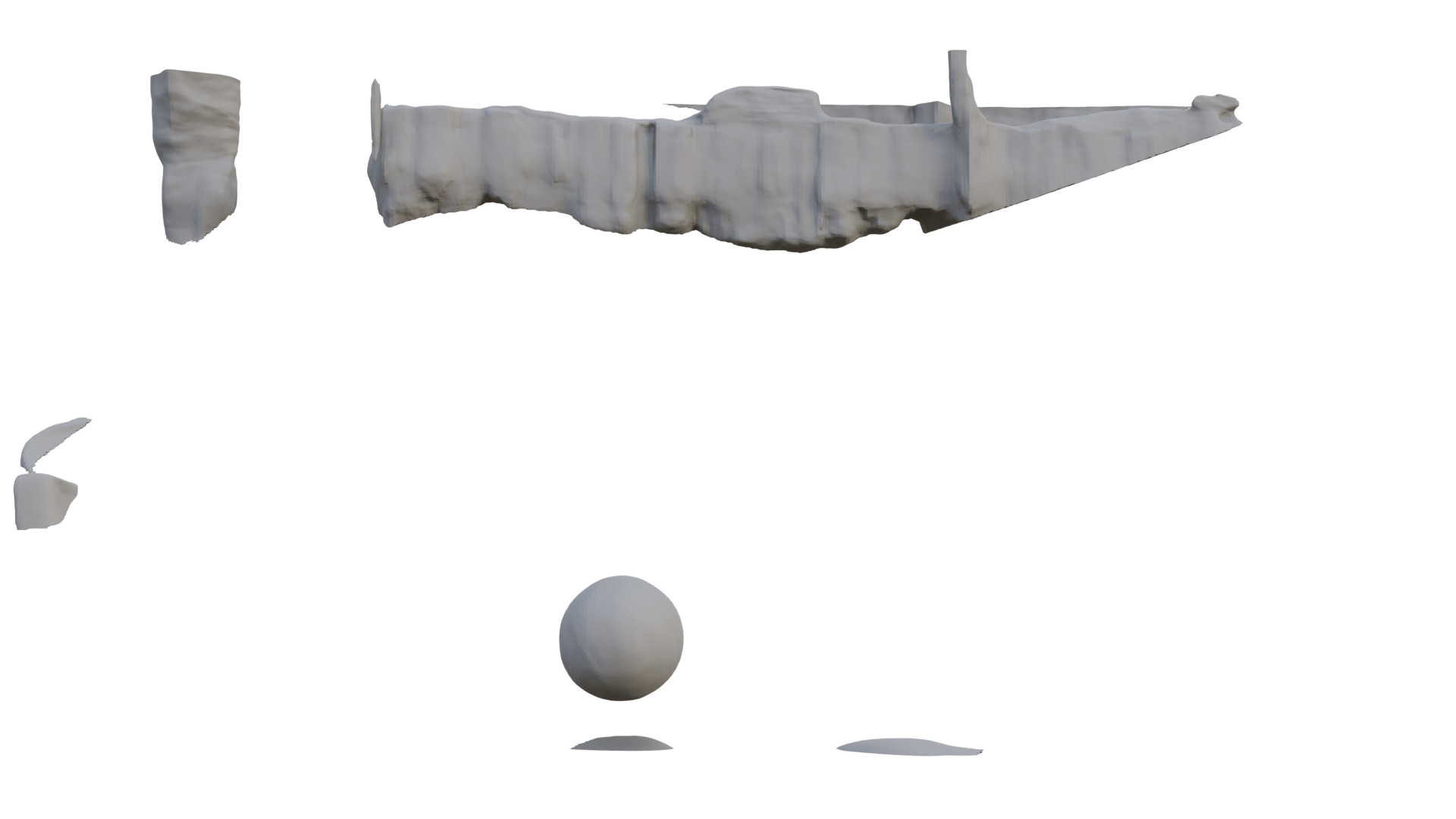} &
    \includegraphics[width=0.32\linewidth]{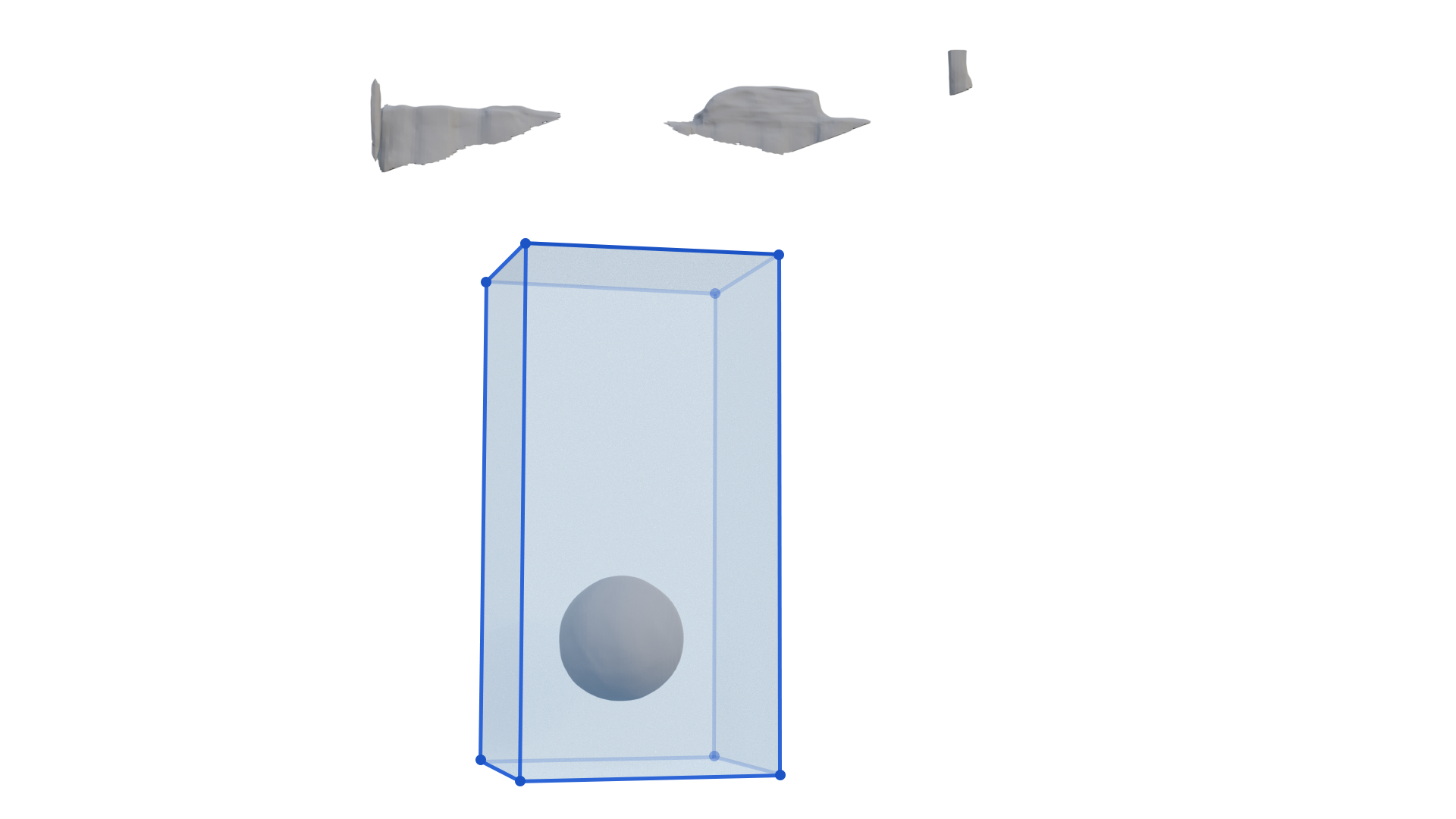}\\
   \small (a) Motion Magnitude $>0.008$   & \small (c) Rigidity $>0.15$   & \small (e) Rigidity $>0.2$ + Bounding Box   \\
    \includegraphics[width=0.32\linewidth]{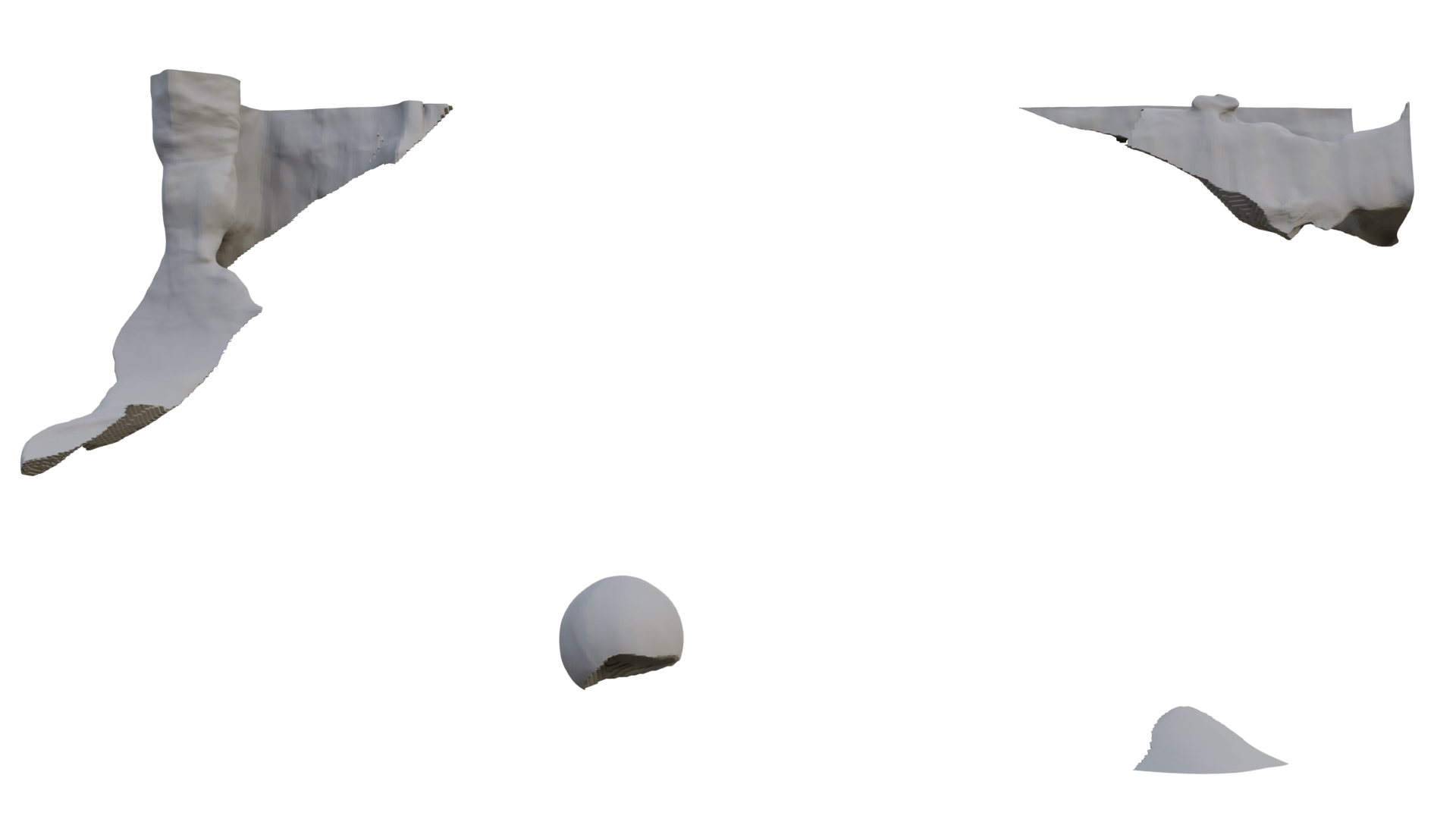} &
    \includegraphics[width=0.32\linewidth]{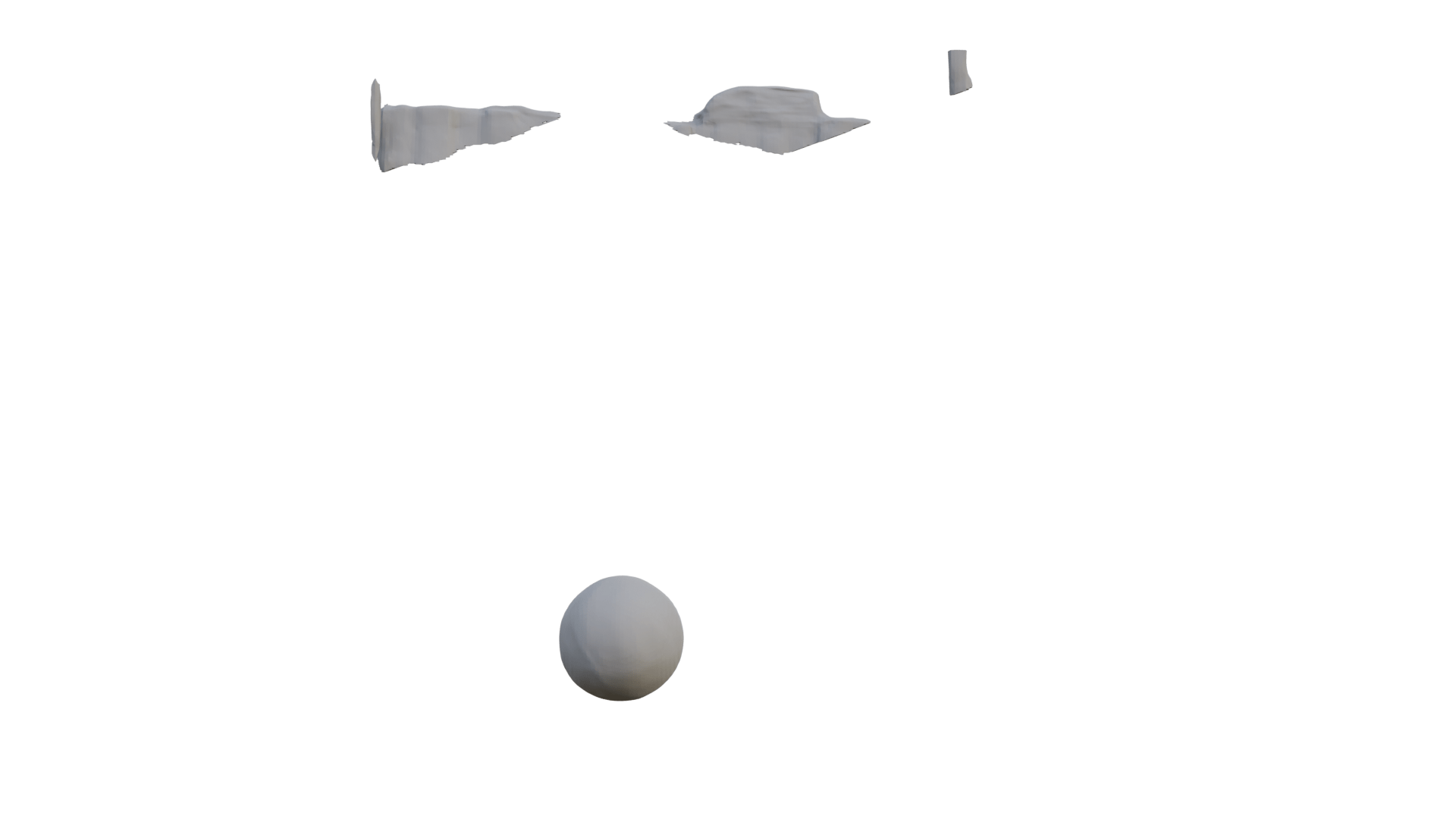} &
    \includegraphics[width=0.32\linewidth]{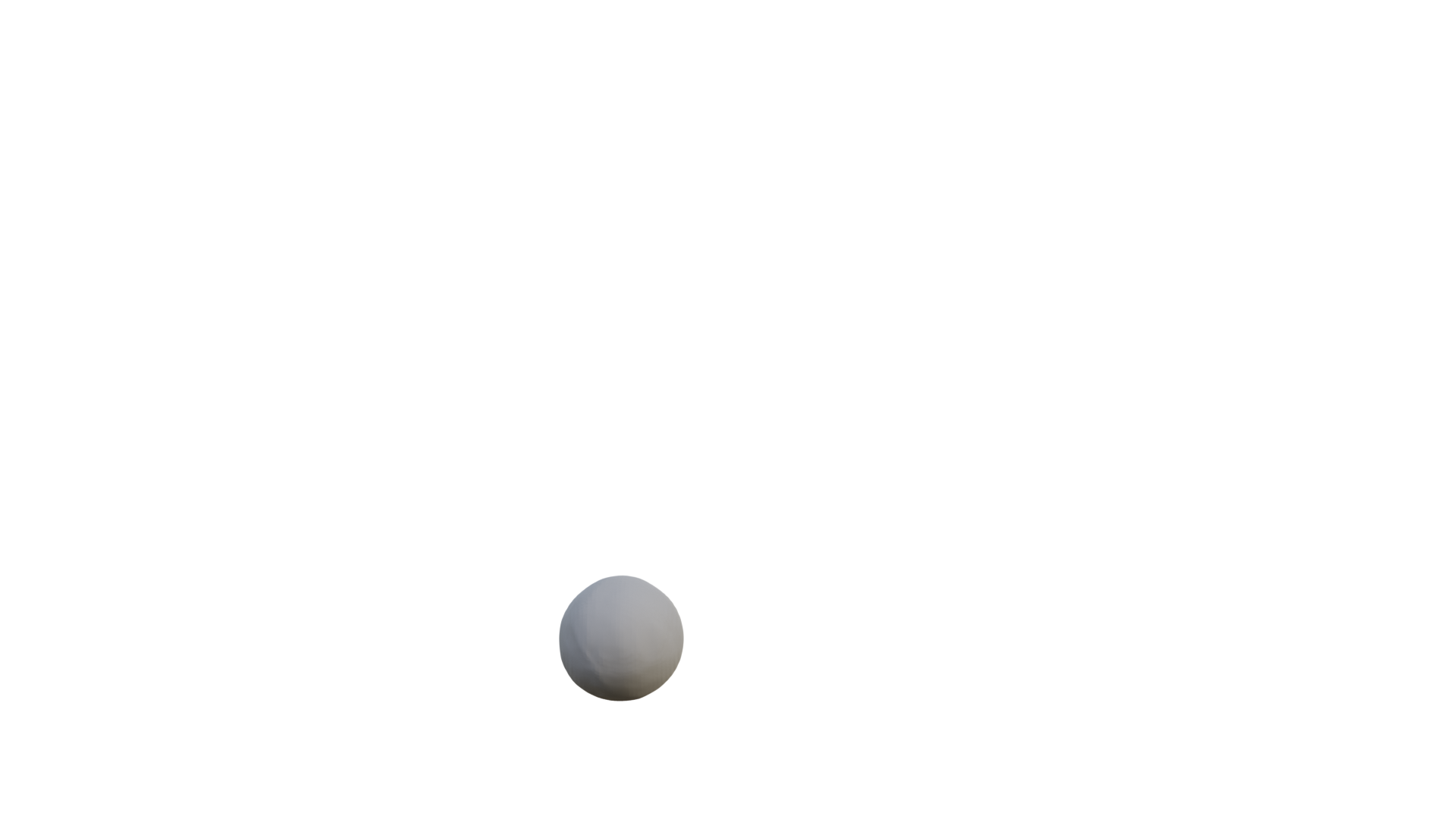} \\
   \small (b) Motion Magnitude $>0.01$  & \small (d) Rigidity $>0.2$   & \small (f) Final Foreground  
\end{tabular}
}
\vspace{-0.5em}
\caption{{\bf Region of Interest.} Motion magnitude (a, b) is an intuitive criterion for static/dynamic separation, but does not work well in practice. The rigidity map (c, d) provides a rough, but reasonable separation. (e, f) We can also specify a more precise ROI with a bounding box.
}
\label{fig:region}
\vspace{-1.5mm}
\end{wrapfigure}

\nps{
\mypara{Rigidity Map.} Finding the correct dynamic foreground object is a prerequisite for physical simulation. The raw output of the SDF field can only reconstruct a global scene, so the rigidity map is used to segment the dynamic part of the scene (assumed to be the foreground in this case). The rigidity map collects the motion information from all frames and assigns a high rigidity mask value to dynamic regions. Figure~\ref{fig:region} (c, d) visualizes the rough segmentation results from using different rigidity thresholds. We empirically find 0.2 to be a reasonable default threshold for separating the foreground from background. If the rigidity map does not provide clean separation, users may manually specify a bounding box covering the moving object swept volume, and get the final foreground as shown in (e, f). An alternative segmentation approach is to use the magnitude of the motion field as a criterion to separate dynamic foreground, i.e. large motion area corresponds to dynamic objects (and the converse). However, in some frames, dynamic parts might only contain small offsets from the canonical frame. For example, a cyclic bouncing ball could overlap with the canonical position and thus have offset values of zero, even though it should be classified as a dynamic area. Figure~\ref{fig:region} (a) and (b) show results of filtering the scene using motion magnitude. It is not a good strategy, since a large portion of the background remains even as the ball is partially clipped off in (b).
}

\vspace*{0em}
\section{Conclusion}
\label{sec:conclusion}
\vspace*{-1em}

In this paper, we design an end-to-end pipeline that can successfully estimate and enable physically-based editing of geometry, appearance, and physics of a dynamic scene from a single video. Some possible future directions include incorporating semantic priors to further constrain the optimization, and leveraging pre-trained networks~\cite{deng2021depth} as supervision to further regularize training. For example, optical flow / scene flow networks could be used as a motion detection signal, while a depth estimation network could help to initialize the signed distance field. Another useful future goal would be to estimate texture~\cite{xiang2021neutex} and UV maps directly, allowing artists to construct 3D models in standard formats with less effort. Training and inference could also be accelerated dramatically by adopting a multi-resolution hash encoding~\cite{mueller2022instant}.  Finally, we could couple this differentiable framework with other types of dynamics, to support a wider range of objects and scenes, e.g. cloth, tetrahedra mesh~\cite{Qiao2021Differentiable}, or even robots~\cite{heiden2021disect}.

\mypara{Acknowledgements.} This research is supported in part by Dr. Barry Mersky and Capital One E-Nnovate Endowed Professorships, and ARL Cooperative Agreement W911NF2120076.
\\\\\\


\small
\bibliographystyle{plainnat}
\bibliography{paper}

\clearpage

\clearpage


\begin{appendices}
\label{sec:appendix}


\section{Weight Function}
\label{ab:weight}

We follow the NeuS~\cite{wang2021neus} paper to set the weight function $w(t)$ in Equation~\ref{eq:rendering}. First, a density function $\rho(t)$ is defined as:
\begin{equation}
    \rho(t) = \max\biggl(\frac{-\frac{d\Phi_h}{dt}(s(\pp(t)))}{\Phi_h(s(\pp(t)))},\ 0\biggr)
\end{equation}

where $\Phi_h(x)=(1+e^{-hx})^{-1}$ is the Sigmoid function, and $w(t)$ can be constructed as:

\begin{equation}
    w(t)=T(t)\rho(t)\text{, where }T(t)=\exp\biggl(-\int_0^t\rho(u)du\biggr).
\end{equation}

\section{Additional Implementation Details}
The Color Network (MLP) consists of 4 dense layers, each with 256 hidden dimensions, while the SDF Network has 8 dense layers of 256 hidden dimensions, the Motion Network has 5 dense layers of 64 hidden dimensions, and the Rigidity Network has 3 dense layers of 32 hidden dimensions.

Similar to NeRF++~\cite{kaizhang2020}, we use a separate NeRF network to render far-away points outside of a unit-sphere region of interest. The Motion and Rigidity networks are not applied to those regions since we assume the background to be static. If we do not disable these networks on the background, we observe significant jittering on those regions.

We have also tried enabling and disabling viewing direction as an input to the Color Network. It turns out that disabling viewing direction can result in smoother geometry, as it prevents over-fitting to the viewing direction as discussed in NeRF++. However, we also observed that disabling the viewing direction led to incorrect geometries (like holes) in some cases.

\section{Potential Applications}
NeuPhysics enables high-resolution mesh reconstruction from a casual video captured by a smartphone. It would be easy for users to collect their own data to customize virtual body avatars for games and social applications. Accurate facial and body models can also be used in virtual try-on, or even customized suits, furniture, and sports equipment. Moreover, the physically-based, dynamic, photorealistic editing technique could be used in designing, content creation, and broader VR/AR applications.

\section{Other Supplementary Materials}
More dynamic visualization results can be found in our supplementary video; we invite the readers to view these videos for visual comparison. A preliminary version of our code is included for reviewing only. We will also open-source a release version in the future for research reproduction with the paper publication.

\begin{figure}
\centering
\begin{tabular}{@{}c@{\hspace{0.3mm}}c@{\hspace{0.3mm}}c@{\hspace{0.3mm}}c@{\hspace{0.3mm}}c@{\hspace{0.3mm}}c@{\hspace{0.3mm}}@{}}
    \includegraphics[width=0.16\linewidth,trim=0 10cm 0 4cm, clip]{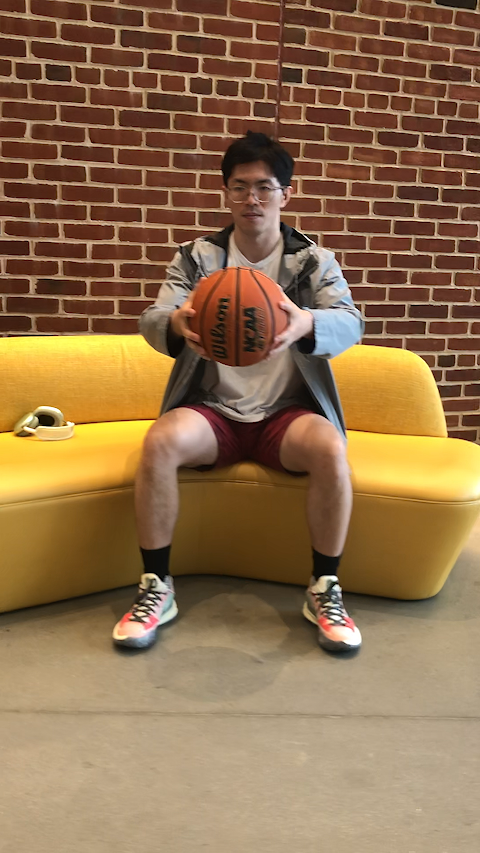} &
    \includegraphics[width=0.16\linewidth,trim=0 10cm 0 4cm, clip]{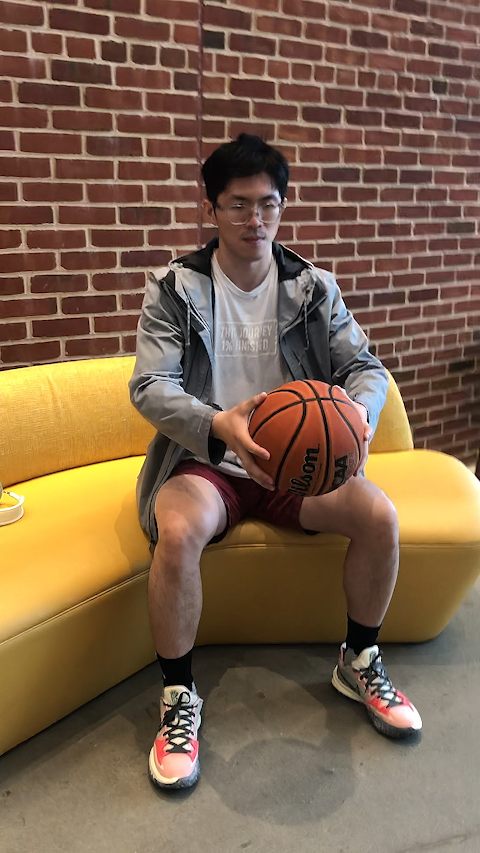} &
    \includegraphics[width=0.16\linewidth,trim=0 10cm 0 4cm, clip]{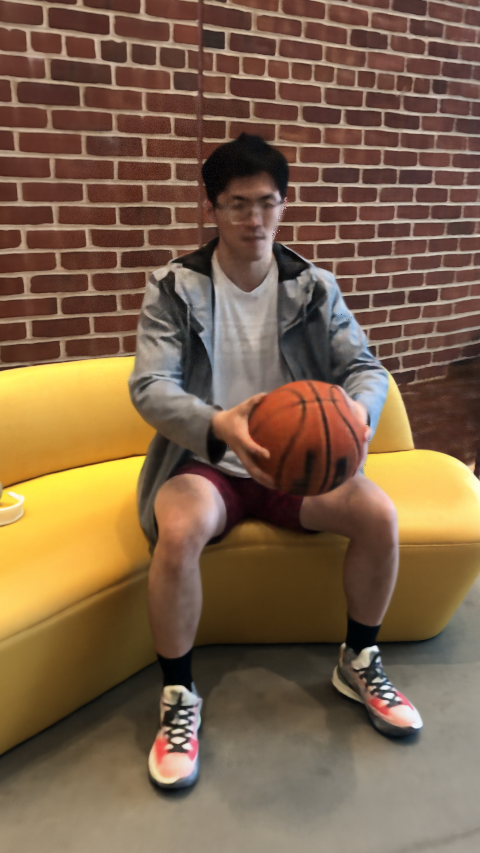} &
    \includegraphics[width=0.16\linewidth,trim=0 10cm 0 4cm, clip]{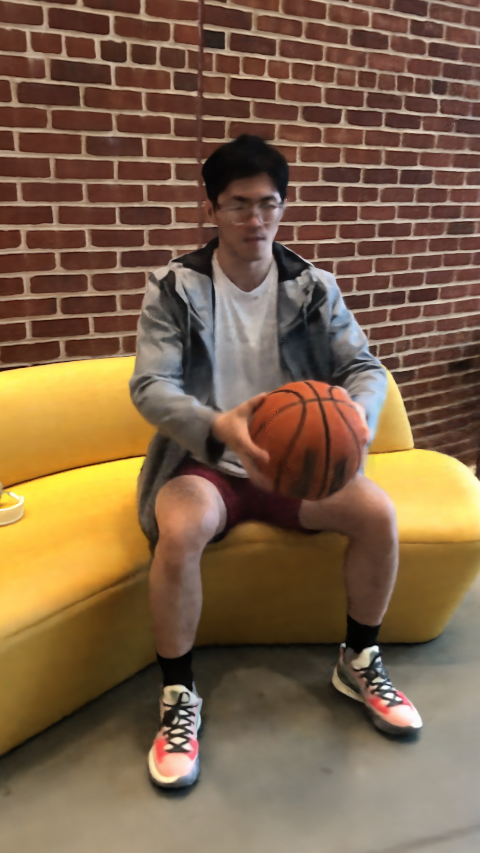} &
    \includegraphics[width=0.16\linewidth,trim=0 10cm 0 4cm, clip]{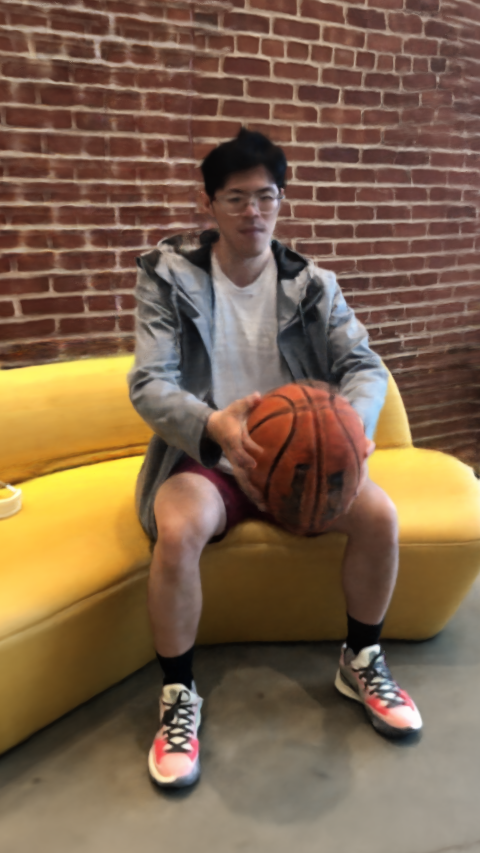} &
    \includegraphics[width=0.16\linewidth,trim=0 10cm 0 4cm, clip]{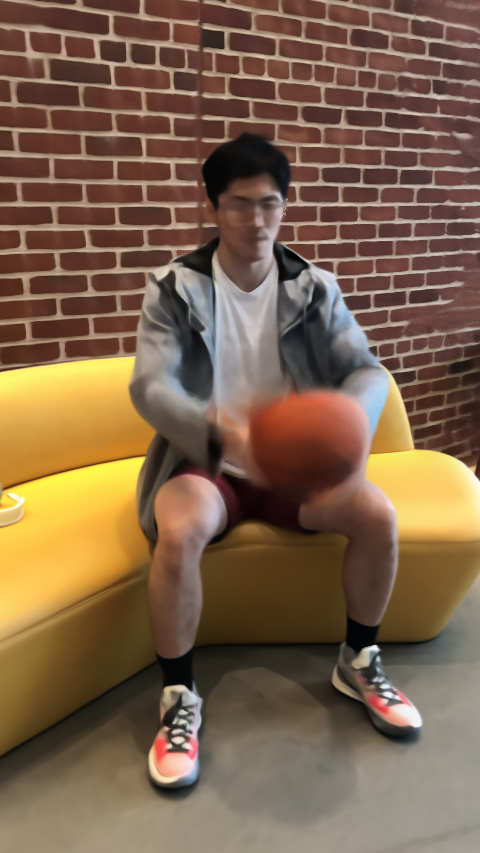} \\
    \includegraphics[width=0.16\linewidth,trim=0 7cm 0 7cm, clip]{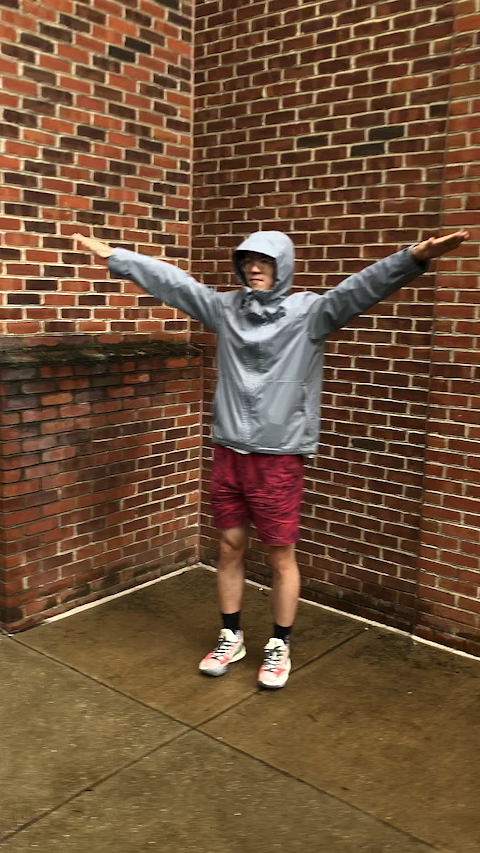} &
    \includegraphics[width=0.16\linewidth,trim=0 7cm 0 7cm, clip]{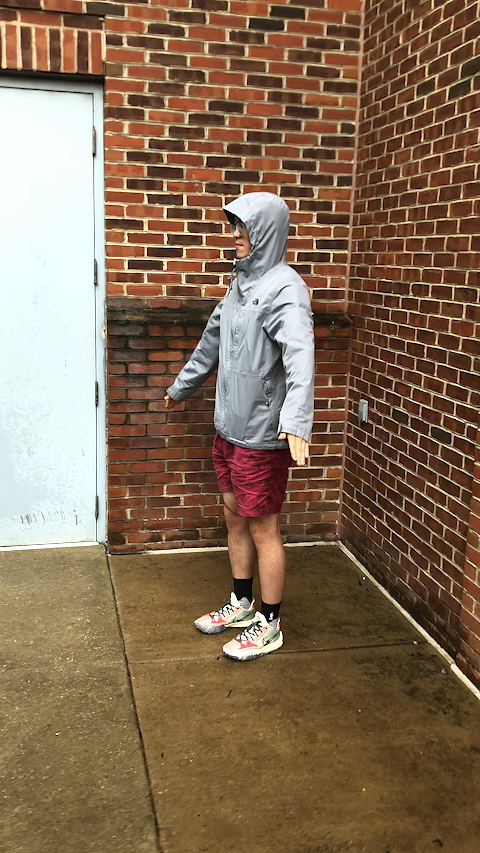} &
    \includegraphics[width=0.16\linewidth,trim=0 7cm 0 7cm, clip]{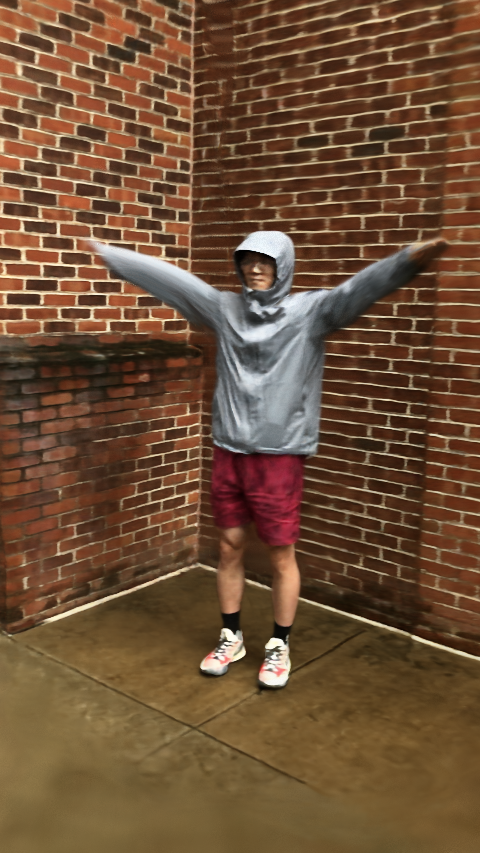} &
    \includegraphics[width=0.16\linewidth,trim=0 7cm 0 7cm, clip]{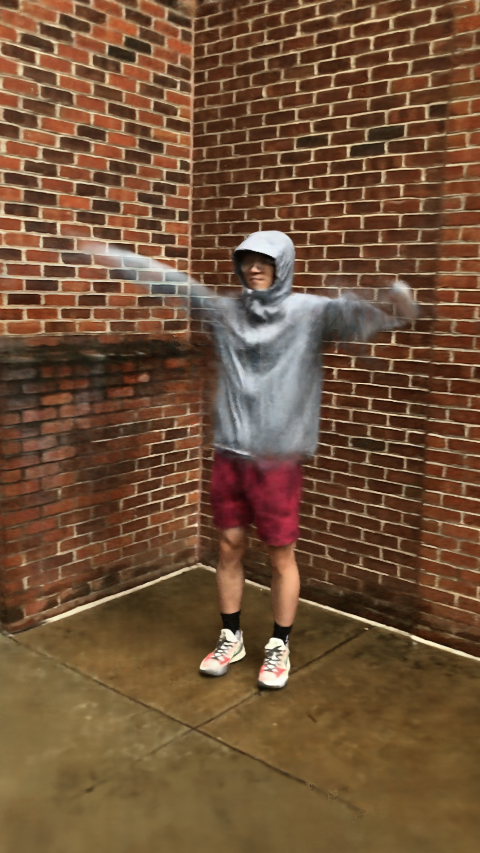} &
    \includegraphics[width=0.16\linewidth,trim=0 7cm 0 7cm, clip]{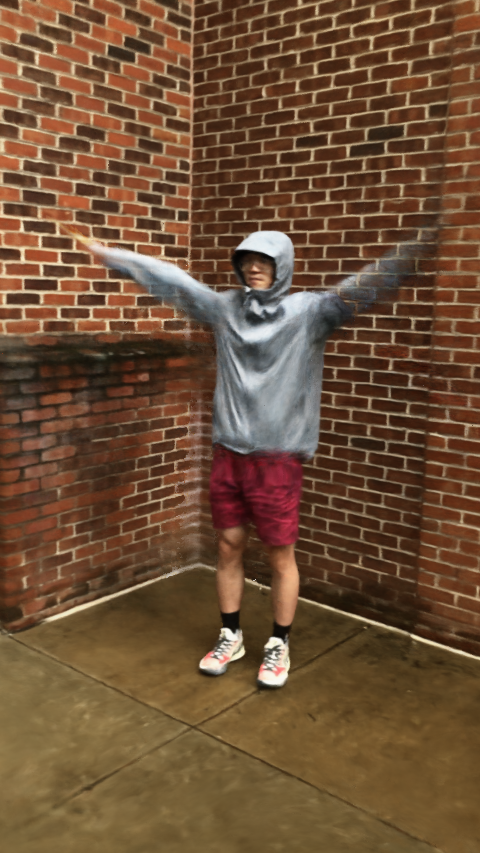} &
    \includegraphics[width=0.16\linewidth,trim=0 7cm 0 7cm, clip]{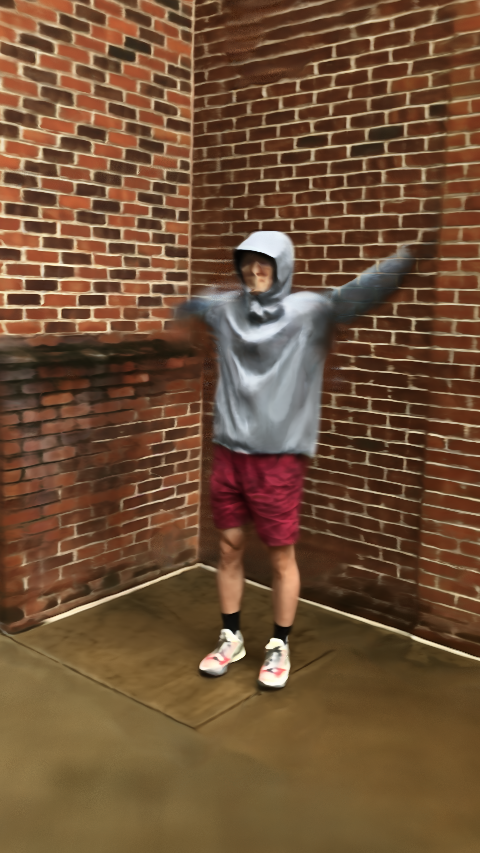} \\
    \includegraphics[width=0.16\linewidth]{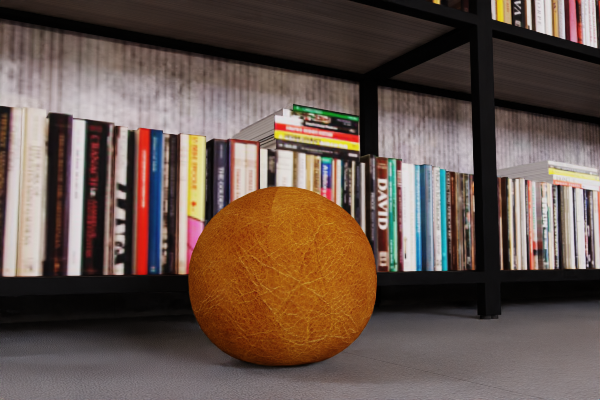} &
    \includegraphics[width=0.16\linewidth]{figures/gt_bm3_0.png} &
    \includegraphics[width=0.16\linewidth]{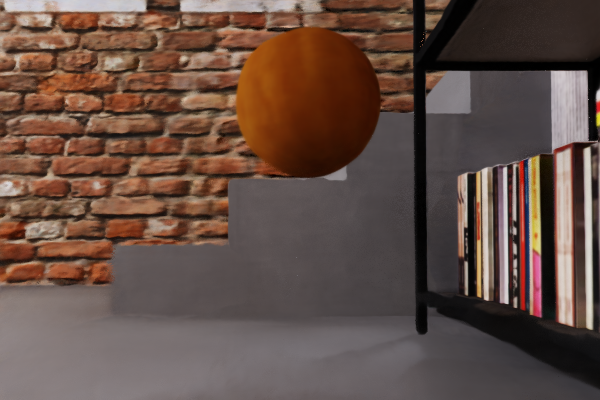} &
    \includegraphics[width=0.16\linewidth]{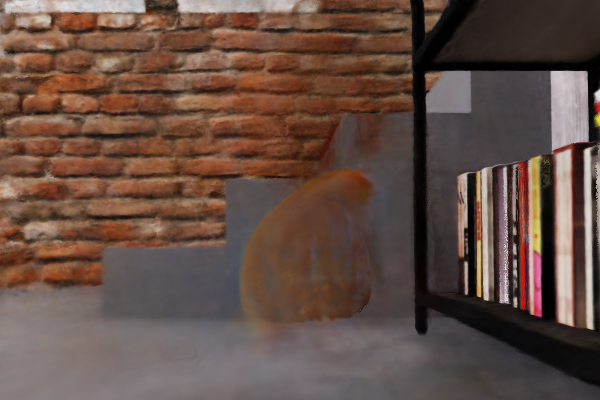} &
    \includegraphics[width=0.16\linewidth]{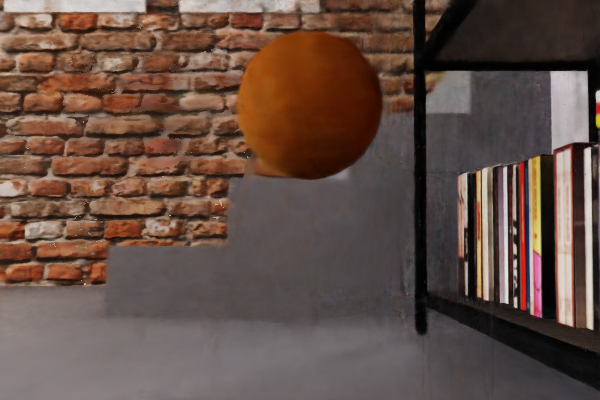} &
    \includegraphics[width=0.16\linewidth]{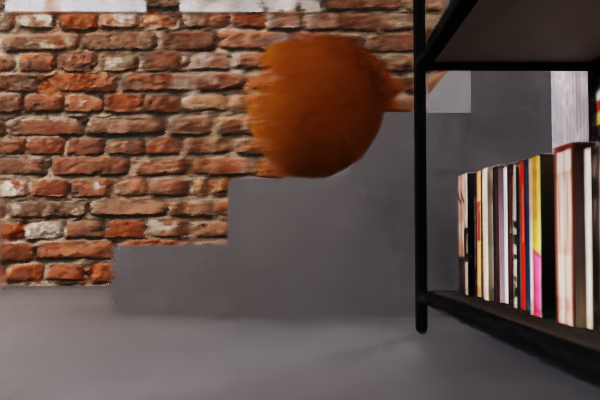} \\
    \includegraphics[width=0.16\linewidth]{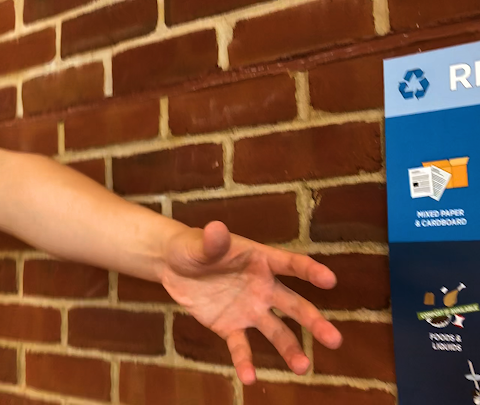} &
    \includegraphics[width=0.16\linewidth]{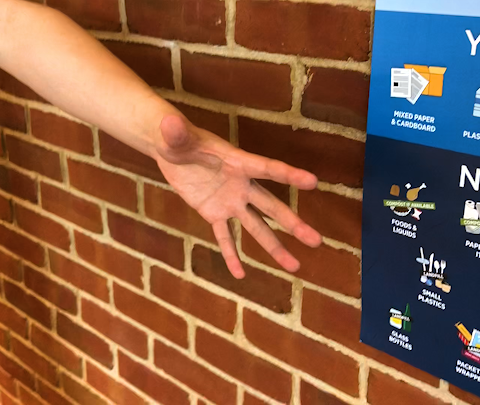} &
    \includegraphics[width=0.16\linewidth]{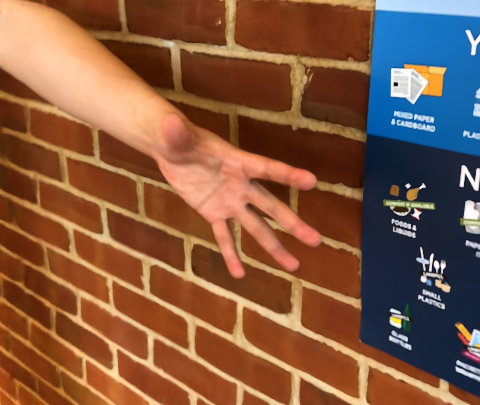} &
    \includegraphics[width=0.16\linewidth]{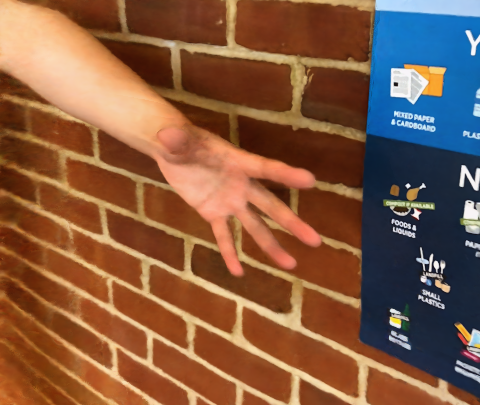} &
    \includegraphics[width=0.16\linewidth]{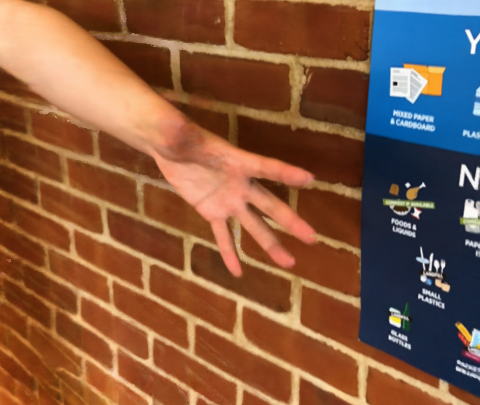} &
    \includegraphics[width=0.16\linewidth]{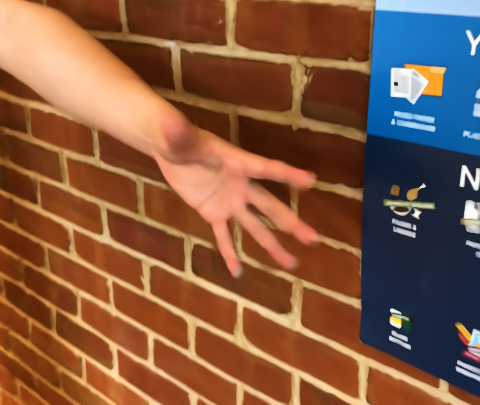} \\
   \small (a) Input 1   & \small (b) Input 2   & \small (c) Ours  & \small (d) NRNeRF~\cite{tretschk2021nonrigid} & \small (e) D-NeRF~\cite{pumarola2020dnerf} & \small (f) NeuS~\cite{wang2021neus} 
\end{tabular}
\vspace{-0.5em}
\caption{{\bf Qualitative evaluation of video reconstruction}. We evaluate four methods on both real and synthetic videos. Our method outperforms other SOTA methods with higher fidelity and less blur.  }
\label{fig:recon}
\vspace{-0.5mm}
\end{figure}

\section{Video Reconstruction}
\label{app:video}
For the video reconstruction task, we evaluate the methods on a dataset consisting of both real-world captured and synthetic videos. The synthetic data are animated by Blender, and the real-world videos are taken by a smartphone in both indoor and outdoor scenes. Figure~\ref{fig:recon} visualizes the datasets as well as the reconstruction videos by different methods. (a, b) are two frames from the input videos where both the scene and the camera positions have changed. (c) is by our method, while (d) NRNeRF~\cite{tretschk2021nonrigid} (e) D-NeRF~\cite{pumarola2020dnerf}, and (f) NeuS~\cite{wang2021neus} are SOTA methods for comparison. Note that (d) and (e) are designed for dynamic videos as input, but not designed for geometry reconstruction. (f) learns SDF field during the video reconstruction as ours does, however, it is originally designed for {\em static} scenes so inconsistency among views would likely undermine its performance. 

\end{appendices}

\end{document}